%% file: ms.tex
\documentclass{article} % For LaTeX2e
 \pdfoutput=1
\usepackage[dvipsnames,table,xcdraw]{xcolor}
\usepackage{iclr2022_conference,times}

\input{math_commands.tex}

\usepackage{epsfig}
\usepackage{graphicx}
\usepackage{amsmath,amsthm}
\usepackage{amssymb}

\usepackage{arydshln}
\usepackage{booktabs}
\usepackage{enumitem}
\usepackage{balance}
\usepackage{combelow}
\usepackage{tabularx}
\usepackage{xspace}
\usepackage{arydshln}
\makeatletter
\def\adl@drawiv#1#2#3{%
        \hskip.5\tabcolsep
        \xleaders#3{#2.5\@tempdimb #1{1}#2.5\@tempdimb}%
                #2\z@ plus1fil minus1fil\relax
        \hskip.5\tabcolsep}
\newcommand{\cdashlinelr}[1]{%
  \noalign{\vskip\aboverulesep
           \global\let\@dashdrawstore\adl@draw
           \global\let\adl@draw\adl@drawiv}
  \cdashline{#1}
  \noalign{\global\let\adl@draw\@dashdrawstore
           \vskip\belowrulesep}}
\makeatother

\makeatletter
\newcommand{\dashrule}[1][black]{%
  \color{#1}\rule[\dimexpr.5ex-.2pt]{4pt}{.4pt}\xleaders\hbox{\rule{4pt}{0pt}\rule[\dimexpr.5ex-.2pt]{4pt}{.4pt}}\hfill\kern0pt%
}
\newcommand{\rulecolor}[1]{%
  \def\CT@arc@{\color{#1}}%
}
\makeatother

\makeatletter
\DeclareRobustCommand\onedot{\futurelet\@let@token\@onedot}
\def\@onedot{\ifx\@let@token.\else.\null\fi\xspace}

\def\eg{\emph{e.g}\onedot} 
\def\ie{\emph{i.e}\onedot\xspace}

\makeatother

\usepackage{wrapfig}
\usepackage[stable]{footmisc}

\usepackage{multirow}
\usepackage{makecell}
\usepackage{mathtools}
\usepackage{algorithm}
\usepackage{algpseudocode}
\usepackage{algorithmicx}
\usepackage{eqparbox,array}

\usepackage{sidecap}
\usepackage[english]{babel}
\newtheorem{theo}{Theorem}
\usepackage[pagebackref=true,breaklinks=true,letterpaper=true,colorlinks,bookmarks=false]{hyperref}

\definecolor{Gray}{gray}{0.9}
\definecolor{azure}{rgb}{0.0, 0.5, 1.0}
\newcommand{\method}{CLIB\xspace}
\newcommand{\methodfull}{Continual Learning for i-Blurry\xspace}
\newcommand{\rev}[1]{{#1}}
\title{Online Continual Learning on Class Incremental Blurry Task Configuration with Anytime Inference}

\makeatletter
\def\thanks#1{\protected@xdef\@thanks{\@thanks
        \protect\footnotetext{#1}}}
\makeatother

\makeatletter
\newcommand{\printfnsymbol}[1]{%
  \textsuperscript{\@fnsymbol{#1}}%
}
\makeatother

\author{Hyunseo Koh\textsuperscript{1,3,*}
\thanks{\hspace{-1.5em}$^*$ indicates equal contribution.}%\protect\phantom{\footnotesize 1}
\hspace{2em}Dahyun Kim\textsuperscript{2,3,*}%\printfnsymbol{1}\printfnsymbol{2}
\hspace{2em}~~~~Jung-Woo Ha\textsuperscript{3}
\hspace{2em}Jonghyun Choi\textsuperscript{3,4,$\dagger$}
\thanks{\hspace{-1.5em}$^\dagger$ indicates corresponding author.}
\thanks{\hspace{-1.5em}This work was done while HK, DK and JC were interns and an AI technical advisor at NAVER AI Lab.}\\
{\textsuperscript{1}GIST, South Korea~~~~~
\textsuperscript{2}Upstage AI Research~~~
\textsuperscript{3}NAVER AI Lab.~~~~~
\textsuperscript{4}Yonsei University}\\
{\tt\small {hyunseo8157@gm.gist.ac.kr, kdahyun@upstage.ai}}\\
{\tt\small {jungwoo.ha@navercorp.com, jc@yonsei.ac.kr}}

}

\iclrfinalcopy % Uncomment for camera-ready version, but NOT for submission.
\begin{document}

\maketitle

\input{contents_camready}

\bibliography{iclr2022_conference}
\bibliographystyle{iclr2022_conference}

\newpage

\appendix
\section{Appendix}
\input{appendix_camready}

\end{document}

%% file: math_commands.tex
\usepackage{amsmath,amsfonts,bm}

\def\eqref#1{equation~\ref{#1}}
\def\1{\bm{1}}

\DeclareMathAlphabet{\mathsfit}{\encodingdefault}{\sfdefault}{m}{sl}
\SetMathAlphabet{\mathsfit}{bold}{\encodingdefault}{\sfdefault}{bx}{n}

\DeclareMathOperator*{\argmax}{arg\,max}
\DeclareMathOperator*{\argmin}{arg\,min}

%% file: contents_camready.tex
\begin{abstract}
Despite rapid advances in continual learning, a large body of research is devoted to improving performance in the existing setups.
While a handful of work do propose new continual learning setups, they still lack practicality in certain aspects.
For better practicality, we first propose a novel continual learning setup that is online, task-free, class-incremental, of blurry task boundaries and subject to inference queries at any moment.
We additionally propose a new metric to better measure the performance of the continual learning methods subject to inference queries at any moment.
To address the challenging setup and evaluation protocol, we propose an effective method that employs a new memory management scheme and novel learning techniques.
Our empirical validation demonstrates that the proposed method outperforms prior arts by large margins. Code and data splits are available at \url{https://github.com/naver-ai/i-Blurry}.
\end{abstract}

%%%%%%%%%%%%%%%%%%%%%%%%%%%%%%%%%%%%%%%%%%%%%%%%%%%%%%%%%%%%%%%%%%%%%%%%%%%%%%%%%%%%%%%%%%%%%%%%%%%%%%%%%%%%%%%%%%
\section{Introduction}
\label{sec:intro}
%Google doc 스토라라인을 참조하여 전면 재수정하면 좋을 것 같음.
% 1. CL 정의 및 중요성 한줄요약
% 2. We argue the limitations of current CL problem definition wrt three criteria
% 3. We design more practical and challenging CL setup by making three criteria more realistic. iBlurry.
% 4. To address this setup, we propose a baseline method using components and a new metric considering our iBlurry setup.
% 5. Current CL methods suffer from our realistic setup. But our models deal with iBlurry setup. More refined and sophisticated method would improve the performances.

Continual learning (CL) is a learning scenario where a model learns from a continuous and online stream of data and is regarded as a more realistic and practical learning setup than offline learning on a fixed dataset~\citep{he20incremental}.
However, many CL methods still focus on the offline setup~\citep{ewc, icarl, saha2021gradient} instead of the more realistic online setup.
These methods assume access to a large storage, storing the entire data of the current task and iterating on it multiple times.
On the other hand, we are interested extensively in the more realistic online setup where only a small memory is allowed as storage.
Meanwhile, even for the online CL methods, we argue they have room for more practical and realistic improvements concerning multiple crucial aspects.
The aspects include the class distributions such as the disjoint~\citep{icarl} or the blurry~\citep{aljundi2019gradient} splits and the evaluation metric that focuses only on the task accuracy such as average task accuracy ($A_{avg}$).

The two main assumptions on the class distributions in existing CL setups, \ie, the disjoint and blurry splits, are less realistic for the following reasons. 
The disjoint split assumes no classes overlap over different tasks; already observed classes will never appear again. 

The above is not plausible because already observed classes can still appear later on in real-world scenarios (see Fig. 2 of \citep{bang2021rainbow}). 
On the other hand, in the blurry split~\citep{aljundi2019gradient} no new classes appear after the first task even though the split assumes overlapping classes over tasks.
This is also not plausible as observing new classes is common in real-world scenarios.
% Therefore, the blurry split that does not introduce new classes is less realistic since the CL model aims to address the scenarios in which new classes can appear as the training progresses.

The typical evaluation metric such as $A_{avg}$ in which the accuracy is measured only at the task transition is also less realistic. 
It implicitly assumes that no inference queries occur in the middle of a task. 
However, in real-world scenarios, inference queries can occur \emph{at any-time}. Moreover, there is no explicit task transition boundary in most real-world scenarios.
Thus, it is desirable for CL models to provide good inference results at any time. % while the training is being done. 
To accurately evaluate whether a CL model is effective at such `any-time' inference, we need a new metric for CL models. % throughout a given task.

% The offline learning, employed by many CL methods~\citep{ewc, icarl, saha2021gradient}, refers to CL methods where models can store the entire data of the current task and iterate on the stored data multiple times.
% However, it assumes large storage while there is only limited storage available to the CL model in the real-world scenarios.
% Thus, CL models cannot store an entire task's data and reuse it multiple times.
% In addition, some CL methods assume access to task descriptors that tells the model when a task transition happens.
% This is also unrealistic as in the real-world, the task transition times are stochastic.

In order to address the issues of the current CL setups, we propose a new CL setup that is more realistic and practical by considering the following criteria:
First, the class distribution is comprised of the advantages from both blurry and disjoint.
That is, we assume that the model continuously encounters new classes as tasks continue, \ie, class-incremental and that classes overlap across tasks, \ie, blurry task boundaries, while not suffering from the restrictions of blurry and disjoint.
Second, the model is evaluated throughout training and inference such that it can be evaluated for any-time inference. 
% Finally, the model needs to learn in an online manner where only a small-sized memory is allowed to be used to store a continuous data stream with no access to task descriptors, \ie, task-free.
We call this new continual learning setup {\bf `i-Blurry'}.

For the i-Blurry setup, we first propose a plausible baseline using experience replay (ER) with reservoir sampling and a tuned learning rate scheduling.
While existing online CL methods are applicable to the i-Blurry setup, they perform only marginally better than our baseline or often worse.

To better handle the i-Blurry setup, we propose a novel continual learning method, which improves the baseline in three aspects.
We design a new memory management scheme to discard samples using a per-sample importance score that reflects how useful a sample is for training.
We then propose to draw training samples only from the memory instead of drawing them from both memory and the online stream as is done in ER.
Finally, we propose a new learning rate scheduling to adaptively decide whether to increase or decrease the learning rate based on the loss trajectory, \ie a data-driven manner. % which is more suitable for the i-Blurry setup.
% Then, we propose a new data distribution setting called `i-blurry' in which the classes are no longer disjoint with new classes also appearing after the first task.
To evaluate the algorithms in the new setup, we evaluate methods by conventional metrics, and further define a new metric called `area under the curve of accuracy' ($A_\text{AUC}$) which measures  the model's accuracy throughout training.
% Our code will be released \href{}{here}.

We summarize our contributions as follows:
\vspace{-0.5em}
\begin{itemize}[leftmargin=10pt]
\setlength\itemsep{-0.2em}
    \item Proposing a new CL setup called i-Blurry, which addresses a more realistic setting that is online, task-free, class-incremental, of blurry task boundaries, and subject to any-time inference.
    \item Proposing a novel online and task-free CL method by a new memory management, memory usage, and learning rate scheduling strategy.
    \item Outperforming existing CL models by large margins on multiple datasets and settings.
    \item Proposing a new metric to better measure a CL model's capability for the desirable any-time inference.
\end{itemize}

%%%%%%%%%%%%%%%%%%%%%%%%%%%%%%%%%%%%%%%%%%%%%%%%%%%%%%%%%
\section{Related Work}
\label{sec:related}

\textbf{Continual learning setups.}
There are many CL setups that have been proposed to reflect the real-world scenario of training a learning model from a stream of data~\citep{Prabhu2020GDumbAS}.
%; using different constraints, difficulty, and practicality all varies across different setups~\citep{Prabhu2020GDumbAS}. 
We categorize them in the following aspects for brevity. % of the large literature.

First, we categorize them into (1) task-incremental (\emph{task-IL}) and (2) class-incremental learning (\emph{class-IL}), depending on whether the task-ID is given at test time. 
Task-IL, also called multi-head setup, assumes that task-ID is given at test time~\citep{LopezPaz2017GradientEM,Aljundi2018MemoryAS, AGEM}. 
In contrast, in class-IL, or single-head setup, task-ID is not given at test time and has to be inferred~\citep{icarl, bic, aljundi2019online}.
Class-IL is more challenging than task-IL, but is also more realistic since task-ID will not likely be given in the real-world scenario~\citep{Prabhu2020GDumbAS}.
% We can also consider whether the task ID is given at training time. 
Most CL works assume that task ID is provided at training time, allowing CL methods to utilize the task ID to save model parameters at task boundaries~\citep{ewc, rwalk} for later use. 
However, this assumption is impractical~\citep{lee2019neural} since real-world data usually do not have clear task boundaries.
To address this issue, a task-free setup~\citep{aljundi2019task}, where task-ID at training is not available, has been proposed.
We focus extensively on the task-free setup as it is challenging and being actively investigated recently~\citep{kim20PRS, lee2019neural, aljundi2019gradient}.

We now categorize CL setups into \emph{disjoint} and \emph{blurry} setup by how the data split is configured.
In the disjoint task setup, each task consists of a set of classes disjoint from all other tasks.
But the disjoint setup is less realistic as the classes in the real-world can appear at any time not only in a disjoint manner.
Recently, to make the setup more realistic, a blurry task setup has been proposed and investigated~\citep{aljundi2019gradient, Prabhu2020GDumbAS,bang2021rainbow}, where $100-M\%$ of the sampels are from the dominant class of the task and $M\%$ of the samples are from all classes, where $M$ is the blurry level~\citep{aljundi2019gradient}.
However, the blurry setup assumes \emph{no class is added} in new tasks, \ie, not class-incremental, which makes the setup still not quite realistic.

Finally, depending on how many samples are streamed at a time, we categorize CL setups into \emph{online}~\citep{ER, aljundi2019online, AGEM} and \emph{offline}~\citep{bic, icarl, rwalk, castro2018eccv}. 
In the offline setup, all data from the current task can be used an unlimited number of times. 
This is impractical since it requires additional memory of size equal to the current task's data.
For the online setup, there are many notions of `online' that differs in each literature.
\cite{Prabhu2020GDumbAS, bang2021rainbow} refer online to a setup using each streamed sample \emph{only once} to train a model while \cite{aljundi2019gradient, aljundi2019online} refer online to a setup where only one or a few samples are streamed at a time. 
We follow the latter as the former allows storing the whole task's data, which is similar to offline and less realistic.
%samples being streamed in units of batch, whereas in an offline setup samples are streamed in units of task. 
%According to this definition, some methods such as RM~\citep{bang2021rainbow} which trains for one epoch but uses whole task data at the same time are not considered online in this paper.

In this work, we propose a novel CL setup that is online, task-free, class-incremental, of blurry task boundaries, and subject to any-time inference as the most realistic setup for continual learning.
Note that task-free and class-incremental are compatible~\citep{gmed, van2021class}.

%--------------------------------------
\textbf{Continual learning methods.}
Given neural networks would suffer from catastrophic forgetting~\citep{mccloskeyC89,ratcliff90}, the online nature of streaming data in continual learning generally aggravates the issue.
To alleviate the forgetting, there are various proposals to store the previous task information; (1) regularization, (2) replay, and (3) parameter isolation.

(1) Regularization methods~\citep{ewc, Zenke2017ContinualLT, Lee2017OvercomingCF, ebrahimi20uncertainty} store previous task information in the form of model priors and use it for regularizing the neural network currently being trained.
(2) Replay methods store a subset of the samples from the previous tasks in an \emph{episodic memory}~\citep{icarl, castro2018eccv, AGEM, bic, kimmer} or keep a generative model that is trained to generate previous task samples~\citep{Shin2017ContinualLW,Wu2018MemoryRG,hu2018overcoming,Cong2020GANMW}.
The sampled or generated examplars are replayed on future tasks and used for distillation, constrained training, or joint training.
(3) Parameter isolation methods augment the networks~\citep{Rusu2016ProgressiveNN, Lee2017LifelongLW, aljundi17expert} or decompose the network into subnetworks for each task~\citep{mallya17packnet, cheung19superposition, yoon2020apd}.

Since (1), (2), and (3) all utilize different ways of storing information that incurs parameter storage costs, episodic memory requirement and increase in network size respectively, a fair comparison among the methods is not straighforward. 
We mostly compare our method with episodic memory-based methods~\citep{aljundi2019online, Prabhu2020GDumbAS, bang2021rainbow} due to performance, but also with methods that use regularization as well~\citep{rwalk, bic}.

%------------------------------------------
\textbf{Online continual learning.}
Despite being more realistic~\citep{losing18incremental,he20incremental}, online CL have not been popular~\citep{Prabhu2020GDumbAS} due to the difficulty and subtle differences in the setups in the literature. 
ER~\citep{ER} is a simple yet strong episodic memory-based online CL method using reservoir sampling for memory management and jointly trains a model with half of the batch sampled from memory. 
Many online CL methods are based on ER~\citep{aljundi2019gradient,aljundi2019online}. 
GSS~\citep{aljundi2019gradient} selects samples based on cosine similarity of gradients. 
MIR~\citep{aljundi2019online} retrieves maximally interfering samples from memory to use for training.

Different from ER, A-GEM~\citep{AGEM} uses the memory to enforce constraints on the loss trajectory of the stored samples.
GDumb~\citep{Prabhu2020GDumbAS} only updates the memory during training phase and trains from scratch at the test time only using the memory.

The recently proposed RM~\citep{bang2021rainbow} uses an uncertainty-based memory sampling and two-stage training scheme where the model trains for one epoch on the streamed samples and trains extensively only using the memory at the end of each task, delaying most of the learning to the end of each task.
The uncertainty-based memory sampling is not well-suited for online CL and the two-stage training leads to poor any-time inference.
Our method outperforms all online CL methods introduced in this section while strictly adhering to the online and task-free restrictions.
%==================

%%%%%%%%%%%%%%%%%%%%%%%%%%%%%%%%%%%%%%%%%%%%%%%%%%%
%%%%%%%%%%%%%%%%%%%%%%%%%%%%%%%%%%%%%%%%%%%%%%%%%%%%%%%%%%%%%%%
\section{Proposed Continual Learning Setup: i-Blurry}
\label{sec:iBlurry}
For a more realistic and practical CL setup, considering real-world scenarios, we strictly adhere to the online and task-free CL setup~\citep{lee2019neural, losing18incremental}. 
Specifically, we propose a novel CL setup (named as \textbf{i-Blurry}), with two characteristics: 1) class distribution being class incremental and having blurry task boundaries and 2) allowing for any-time inference.

\begin{figure}[t!]
    \centering
    \resizebox{0.9\linewidth}{!}{
    \includegraphics{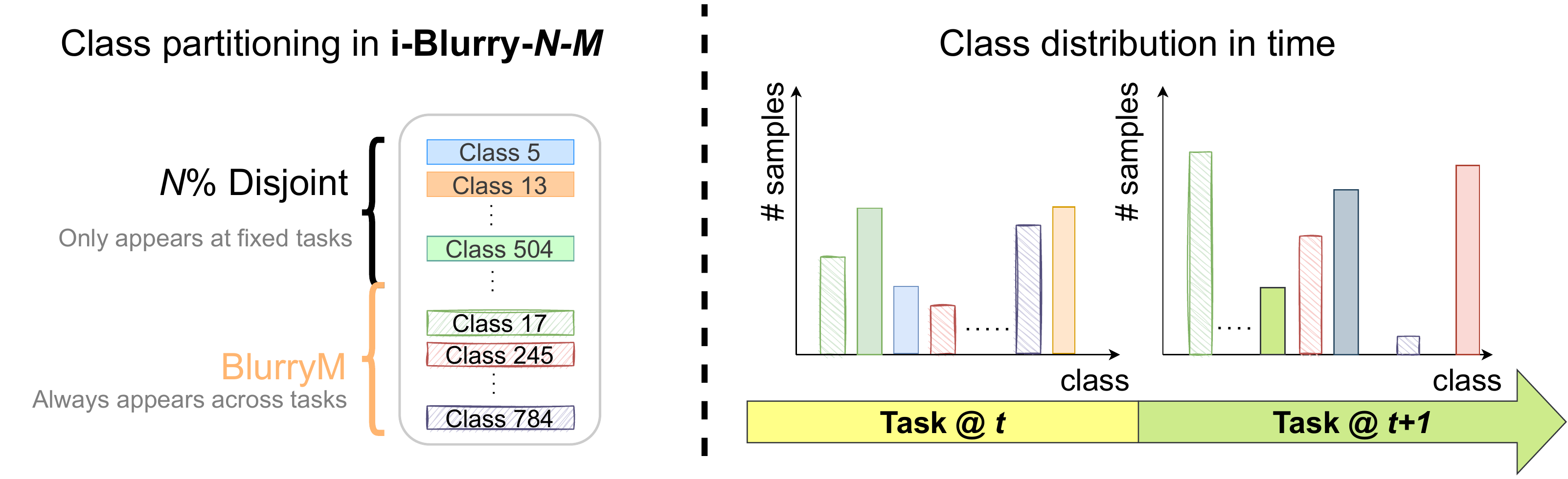}
    }
    \vspace{-1em}
    \caption{i-Blurry-$N$-$M$ split. $N\%$ of classes are partitioned into the disjoint set and the rest into the Blurry$M$ set where $M$ denotes the blurry level~\citep{aljundi2019gradient}.
    To form the i-Blurry-$N$-$M$ task splits, we draw training samples from a uniform distribution from the `disjoint' or the `Blurry$M$' set~\citep{aljundi2019gradient}. The `blurry' classes always appear over the tasks while disjoint classes gradually appear.}
    \vspace{-1em}
    \label{fig:iblurry_split}
\end{figure}

%----------------------------------
\textbf{i-Blurry-N-M Split.} %: i-blurry Split}
\label{sec:iblurry_split}
%The overview of how the i-blurry split is constructed is given in Fig.~\ref{fig:iblurry_split}.
We partition the classes into groups where $N\%$ of the classes are for disjoint and the rest of $100-N\%$ of the classes are used for Blurry$M$ sampling~\citep{aljundi2019gradient}, where $M$ is the blurry level.
Once we determine the partition, we draw samples from the partitioned groups. 
We call the resulting sequence of tasks as \textbf{i-Blurry-$N$-$M$} split.
The i-Blurry-$N$-$M$ splits feature both class-incremental and blurry task boundaries.
Note that the i-Blurry-$N$-$M$ splits generalize previous CL setups.
For instance, $N=100$ is the disjoint split as there are no blurry classes.
$N=0$ is the Blurry$M$ split~\citep{aljundi2019gradient} as there are no disjoint classes.
$M=0$ is the disjoint split as the blurry level is $M=0$~\citep{bang2021rainbow}.
We use multiple i-Blurry-$N$-$M$ splits for reliable empirical validations and share the splits.
Fig.~\ref{fig:iblurry_split} illustrates the i-Blurry-$N$-$M$ split.
\begin{figure}[t!]
    \centering
    \resizebox{1.0\linewidth}{!}{
    \includegraphics{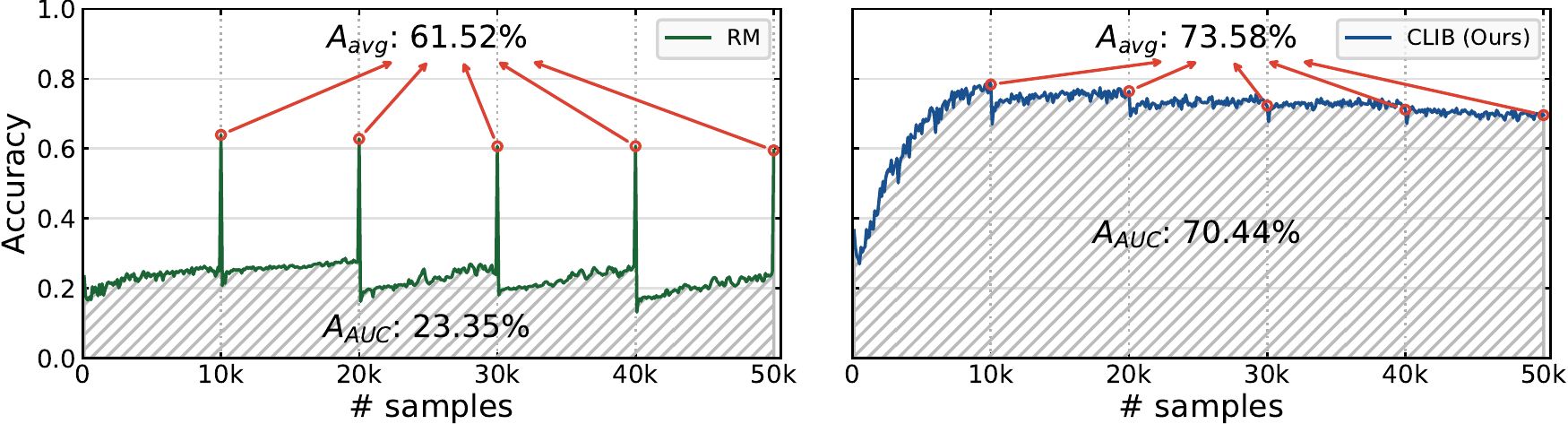}
    }\\
    {\footnotesize \hspace{-5em}(a) Rainbow Memory (RM)~\citep{bang2021rainbow}\hspace{10em} (b) \method \hfill}\\
    \vspace{-0.5em}
    \caption{%Example where $A_\text{AUC}$ is more useful compared to just $A_{avg}$. 
    Comparison of $A_\text{AUC}$ with $A_{avg}$.
    (a) online version of RM~\citep{bang2021rainbow} (b) proposed \method.
    The two-stage method delays most of the training to the end of the task.
    The accuracy-to-\{\# of samples\} plot shows that our method is more effective at any time inference than the two-stage method.
    The difference in $A_{avg}$ for the two methods is much smaller than that of in $A_\text{AUC}$, implying  that $A_\text{AUC}$ captures the effectiveness at any-time inference better.} 
    \vspace{-0.5em}
    \label{fig:gacc}
\end{figure}

%----------------------------------
\textbf{A New Metric -- Area Under the Curve of Accuracy ($A_\text{AUC}$).}
\label{sec:gacc}
Average accuracy (\ie, $A_{avg} ={1\over T} \sum_{i=1}^{T} A_i$ where $A_i$ is the accuracy at the end of the i\textsuperscript{th} task) is one of the widely used measures in continual learning.
% We argue that the widely used $A_{avg} ={1\over T} \sum_{i=1}^{T} A_i$ where $A_i$ is the accuracy at the end of the i\textsuperscript{th} task, has limitations as it measures the accuracy only when the task ends.
But $A_{avg}$ only tells us how good a CL model is at the few discrete moments of task transitions ($5-10$ times for most CL setups) when the model could be queried at any time.
Thus, a CL method could be poor at any-time inference but the $A_{avg}$ may be insufficient to deduce that conclusion due to its temporal sparsity of measurement.
For example, Fig.~\ref{fig:gacc} compares the online version of RM~\citep{bang2021rainbow}, which conducts most of the training by iterating over the memory at the end of a task, with our method.
RM is shown as it is a very recent method that performs well on $A_{avg}$ but particularly poor at any-time inference.
Only evaluating with $A_{avg}$ might give the false sense that the difference between the two methods is not severe.
However, the accuracy-to-\{\# of samples\} curve reveals that our method shows much more consistently high accuracy during training, implying that our method is more suitable for any-time inference than RM.
To alleviate the limitations of $A_{avg}$, we shorten the accuracy measuring frequency to after every $\Delta n$ samples are observed instead of at discrete task transitions.
The new metric is equivalent to the area under the curve (AUC) of the accuracy-to-\{\# of samples\} curve for CL methods when $\Delta n=1$\rev{.}
We call it area under the curve of accuracy ($A_\text{AUC}$):
\begin{equation}
  A_\text{AUC} = \displaystyle\sum_{i=1}^{k} f(i\cdot \Delta n) \cdot \Delta n,
\end{equation}
where the step size $\Delta n$ is the number of samples observed between inference queries and $f(\cdot)$ is the curve in the accuracy-to-\{\# of samples\} plot.
High $A_\text{AUC}$ corresponds to a CL method that consistently maintains high accuracy throughout training.

The large difference in $A_\text{AUC}$ (see Fig.~\ref{fig:iblurry_plot}) implies that delaying strategies like the two-stage training scheme are not effective for any-time inference, a conclusion harder to deduce with just $A_{avg}$.

%-------------------------------------------------------
% \textbf{Online and Task-Free CL.}
% \label{sec:online}
% We reiterate two restrictions that need to be satisfied for online and task-free CL.
% The first restriction is on storage.
% Instead of assuming a very large storage that can store the entire task's data and the ability to reuse the stored data many times as is often done in offline CL~\citep{bic, icarl, ewc}, we follow a more restricted setting where only up to $32$ data points can be stored and reused only up to $3$ times.
% The second restriction is on the access to task descriptors.
% We assume that CL models will not have access to task descriptors, making the setup task-free.
% We believe this is reasonable as we would not know when the data distribution changes in the real-life scenario.

%=========================
\section{Method}
\label{sec:approach}

%-----------------------------
\subsection{A Baseline for i-Blurry Setup}
\label{sec:baseline}
To address the realistic i-Blurry setup, we establish a baseline for the challenging online and task-free i-Blurry setup.
For the memory management policy, we use reservoir sampling~\citep{reservoir} \rev{and for memory usage, we we use experience replay (ER).} 
% Since ER is a simple yet effective method that shows promising performance, we utilize ER in our new baseline.
For the LR scheduling, \rev{we use an exponential LR schedule but reset the LR when a new class is encountered.
Please see Sec.~\ref{sec:baseline} for details.}

\begin{figure}
    \centering
    \resizebox{1\linewidth}{!}{
    \includegraphics{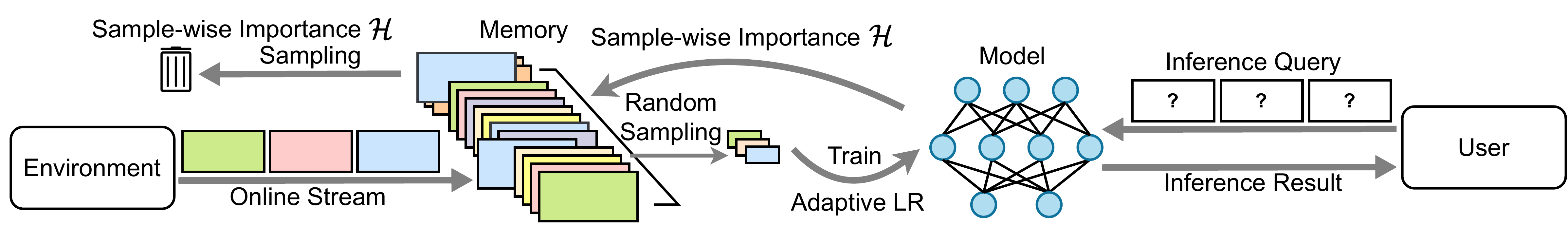}
    }
    \caption{Overview of the proposed \method. We compute sample-wise importance during training to manage our memory. Note that we only draw training samples from the memory whereas ER based methods draw them from both the memory and the online stream.}
    \vspace{-0.5em}
    \label{fig:overview}
\end{figure}

Note that the above baseline still has room to improve.
The reservoir sampling does not consider whether one sample could be more useful for training than the others. %, a key idea in curriculum learning~\citep{bengio2009CL}.
ER uses samples from the stream directly, which can skew the training of CL models to recently observed samples.
While the exponential with reset does increase the LR periodically, the sudden changes may disrupt the CL model.
Thus, we discuss how we can improve the baseline in the following sections.
% \vspace*{-0.2cm}
% \begin{enumerate}[leftmargin=13pt]
%     \item \textbf{First improvement:} blah
%     \item \textbf{Second improvement:} blah
% \end{enumerate}
% \vspace*{-0.15cm}
The final method with all the improvements is illustrated in Fig.~\ref{fig:overview}.
%Thus, for our new baseline we use ER with exponential with reset LR scheduling.

% We improve upon our new baseline from three aspects; the memory management, the memory usage, and the LR schedule.
% An overview of our method is given in Fig.~\ref{fig:overview}

%-----------------------------
\subsection{Sample-wise Importance Based Memory Management}
\label{sec:per_sample}
In reservoir sampling, the samples are removed from the memory at random.
\rev{Inspired by works on sample importance~\citep{kloek1978bayesian, lecun1990optimal, chang2017active, katharopoulos2018not, csiba2018importance}}, we propose a \rev{novel  sampling strategy specific for CL} that removes samples from the memory based on sample-wise importance as following Theorem~\ref{th:per_sample}.
% \begin{itemize}
% \setlength\itemsep{-0.3em}
%   \item Keep the samples that are easy to forget when \emph{not used} for training
%   \item Keep the samples that prevents forgetting of other samples when \emph{used} for training
% \end{itemize}
% They are measured and updated for each sample in memory after every model update using Alg.~\ref{algo:loss_history}. The memory is updated using the two measured criteria, as shown in Alg.~\ref{algo:memory_update}.

\begin{theo}
Let $\mathcal{C}=\mathcal{M}\cup\{(x_{new},y_{new})\}$, $\mathcal{M}$ be a memory from the previous time step, $(x_{new},y_{new})$ be the newly encountered sample, $l(\cdot)$ be the loss and $\theta$ be the model parameters.
Assuming that the model trained with the optimal memory, $\mathcal{M}^{*}$ will induce maximal loss decrease on $\mathcal{C}$, the optimal memory is given by $\mathcal{M}^{*} = \mathcal{C}\setminus\{(\bar{x},\bar{y})\}$ with
\begin{equation}
(\bar{x},\bar{y})=\argmin_{(x_i,y_i)\in\mathcal{C}} \mathbb{E}_\theta\left[\sum_{(x, y)\in\mathcal{C}}l(x, y;\,\theta) - l\left(x, y;\, \theta-\nabla_\theta l(x_i, y_i;\,\theta)\right)\right].
\label{eq:per_sample_crit}
\end{equation}
\begin{proof}
Please see Sec.~\ref{sec:proof_of_th1} for the proof.
\end{proof}
\label{th:per_sample}
\end{theo}
\vspace{-1em}
\rev{Theorem~\ref{th:per_sample} states that when a new sample is appended to the memory and a sample has to be discarded, we should discard the sample that incurs the least loss decrease \ie, least useful for training.} 
We solve Eq.~\ref{eq:per_sample_crit} by keeping track of the sample-wise importance $\mathcal{H}_i$:
\begin{equation}
    \mathcal{H}_i = \mathbb{E}_\theta\left[\sum_{(x, y)\in\mathcal{C}}l(x, y; \,\theta) - l\left(x, y;\, \theta-\nabla_\theta l(x_i, y_i;\,\theta)\right)\right].
\label{eq:import_score}
\end{equation}
Intuitively, $\mathcal{H}$ is the expected loss decrease when the associated sample is used for training. We update $\mathcal{H}$ associated with the samples used for training after every training iteration.
\rev{
Specifically, the model is trained with a random batch from the memory. Then, for all samples in $C$, we measure the loss difference before and after training with the batch. If the loss decreases, the importance scores of the samples in the batch increase and vice versa. Note that because the importance score of a sample is relative to that of other samples in the memory, when the importance scores of the samples in the batch decrease, the scores of the samples not in the batch increase comparatively. Thus, two types of samples have high importance scores; 1) samples that were in the batch when the loss decreased and 2) samples that were not in the batch when the loss increased.}

\rev{The expectation in Eq.~\ref{eq:import_score} is taken over $\theta$'s optimization trajectory \ie, over time. For computational efficiency, we use the exponential moving average as empirical estimates instead.}
The empirical estimates are calculated by the discounted sum of the differences between the actual loss decrease and the predicted loss decrease (see Alg.~\ref{algo:loss_history}).
The memory is updated whenever a new sample is encountered, and the full process is given in Alg.~\ref{algo:memory_update}. 
The memory management strategy significantly outperforms reservoir sampling, especially with memory only training.

%------------------------------------
\subsection{Memory Only Training}
\label{sec:memory_only_training}
ER uses joint training where half of the training batch is obtained from the online stream and the other half from memory. 
\rev{However, we argue that using the streamed samples directly will skew the training to favor the recent samples more.} Thus, we propose to use samples only from the memory for training, without using the streamed samples. 
The memory works as a distribution stabilizer for streamed samples through the memory update process (see Sec.~\ref{sec:per_sample}), and samples are used for training only with the memory. 
We observe that the memory only training improves the performance despite its simple nature.
Note that this is different from~\cite{Prabhu2020GDumbAS} as we train with the memory during the online stream but \cite{Prabhu2020GDumbAS} does not.

%------------------------------------
\subsection{Adaptive Learning Rate Scheduling}
\label{sec:adaptive_lr}
The exponential with reset scheduler resets the LR to the initial value when a new class is encountered.
As the reset occurs regardless of the current LR value, it could result in a large change in LR value.
We argue that such abrupt changes may harm the knowledge learned from previous samples.

Instead, we propose a new data-driven LR scheduling scheme that adaptively \rev{changes} the LR \rev{in a data-driven manner} based on how good the LR is for optimizing over the memory. 

\rev{Specifically, from the current base LR $\bar{\eta}$ and step size $\gamma<1$, we try both a high LR $\bar{\eta}/\gamma$ and a low LR $\bar{\eta}\cdot\gamma$ for training. For each LR, we keep a history of length $m$ that tracks the loss decrease for each LR. When both histories are full, we perform a Student's $t$-test with significance level $\alpha=0.05$ to compare the LRs. If one LR is better, the base LR is set to the better LR, \ie, $\frac{\bar{\eta}}{\gamma}$ or $\bar{\eta}\cdot\gamma$.}

We depict this scheme in Alg.~\ref{algo:lr_schedule} in Sec.~\ref{sec:adaptive_lr_alg}.
\rev{The adaptive LR is important for CL methods because the data distribution changes over time and adapting to the current training data is more useful.}

With all the proposed components, we call our method \textbf{\methodfull} or (\textbf{\method}).

%%%%%%%%%%%%%%%%%%%%%%%%%%%%%%%%%%%%%%%%%%%%%%%%%%%%%%%%%%%%%%%%%%%%%%%%%%%%%%%%%%%%%%%%%%%%%%%%%%%%%%%%%%%%%%%%%%
\section{Experiments}
\label{sec:exp}

\textbf{Experimental Setup.}
We use the CIFAR10, CIFAR100, TinyImageNet, \rev{and ImageNet} datasets for empirical validations.
We use the i-Blurry setup with $N=50$ and $M=10$ (i-Blurry-50-10) for our experiments unless otherwise stated.
All results are averaged over 3 independent runs \rev{except ImageNet~\citep{bic,bang2021rainbow, Prabhu2020GDumbAS}}.
For metrics, we use the average accuracy ($A_{avg}$) and the proposed $A_\text{AUC}$ (see Sec.~\ref{sec:gacc}).
Additional discussion with other metrics such as the forgetting measure ($F_\text{last}$) can be found in Sec.~\ref{sec:app_f_measure}.

\textbf{Implementation Details.}
For all methods, we fix the batch size and the number of updates per streamed samples observed when possible. For CIFAR10, we use a batch size of 16 and 1 updates per streamed sample. When using ER, this translates to 8 updates using the same streamed batch, since each batch contains 8 streamed samples. For CIFAR100, we use a batch size of 16 and 3 updates per streamed sample. For TinyImageNet, we use a batch size of 32 and 3 updates per streamed sample.
\rev{For ImageNet, we use a batch size of 256 and 1 update per every 4 streamed samples.}
We use ResNet-18 as the model for CIFAR10 and ResNet-34 for CIFAR100 and TinyImageNet. For all methods, we apply AutoAugment~\citep{AutoAugment} and CutMix~\citep{yun2019cutmix}  following RM~\citep{bang2021rainbow}. For memory size, we use 500, 2000, 4000, 20000 for CIFAR10, CIFAR100, TinyImageNet, \rev{ImageNet}, respectively. \rev{We use 5 tasks for CIFAR10, CIFAR100, and TinyImageNet and 10 tasks for ImageNet.}
\rev{We follow prior works~\citep{bang2021rainbow,Prabhu2020GDumbAS} for choosing number of updates per sample, memory size, batch size, and number of tasks.}
Additional analysis on the sample memory size \rev{and number of tasks} can be found in Sec.~\ref{sec:app_memory_size}. Adam optimizer with initial LR of 0.0003 is used. Exponential with reset LR schedule is applied for all methods except ours and GDumb, with $\gamma=0.9999$ for CIFAR datasets and $\gamma=0.99995$ for TinyImageNet \rev{and ImageNet}. Ours use adaptive LR with $\gamma=0.95, m=10$ for all datasets, GDumb~\citep{Prabhu2020GDumbAS} and RM~\citep{bang2021rainbow} follow original settings in their respective paper. 
All experiments were performed based on NAVER Smart Machine Learning (NSML) platform~\citep{kim2018nsml,sung2017nsml}.
All code and i-Blurry-$N$-$M$ splits (see Supp.) is at \url{https://github.com/naver-ai/i-Blurry}.

\textbf{Baselines.} 
We compare our method with both online CL methods and ones that can be extended to the online setting; EWC++~\citep{rwalk}, BiC~\citep{bic}, GDumb~\citep{Prabhu2020GDumbAS}, A-GEM~\citep{AGEM}, MIR~\citep{aljundi2019online} and RM~\citep{bang2021rainbow}.
$\cdot^{\dagger}$ indicates that the two-stage training scheme~\citep{bang2021rainbow} is used.
For details of the online versions of these methods, see Sec.~\ref{sec:app_online_detail}.
Note that A-GEM performs particularly worse (also observed in~\citep{Prabhu2020GDumbAS, mai2021online}) as A-GEM was designed for the task-incremental setup and our setting is task-free.
We discuss the comparisons to A-GEM in Sec.~\ref{sec:app_a_gem}.

% Unless otherwise stated in the original method, all methods are implemented using ER, reservoir sampling, and exponential decay with reset LR schedule.
\begin{table}[t!]
    \centering
    \resizebox{0.9\linewidth}{!}{
\begin{tabular}{lcccccccc}
\toprule
\multirow{2}{*}{Methods} & \multicolumn{2}{c}{CIFAR10} & \multicolumn{2}{c}{CIFAR100}            & \multicolumn{2}{c}{TinyImageNet} &
\multicolumn{2}{c}{\rev{ImageNet}}\\ \cmidrule(lr){2-3} \cmidrule(lr){4-5} \cmidrule(lr){6-7} \cmidrule(lr){8-9}
     & $A_\text{AUC}$ & $A_\text{avg}$ & $A_\text{AUC}$ & $A_\text{avg}$ & $A_\text{AUC}$ & $A_\text{avg}$ & \rev{$A_\text{AUC}$} & \rev{$A_\text{avg}$} \\
\cmidrule(lr){1-1} \cmidrule(lr){2-3} \cmidrule(lr){4-5} \cmidrule(lr){6-7} \cmidrule{8-9}
\rev{Joint Training\textsuperscript{$\ddagger$} (Soft Upper Bound)} & \multicolumn{2}{c}{\rev{96.03}} & \multicolumn{2}{c}{\rev{79.89}} & \multicolumn{2}{c}{\rev{53.05}} &
\multicolumn{2}{c}{\rev{69.26}}\\ 
% \midrule
\cmidrule(lr){1-1} \cmidrule(lr){2-3} \cmidrule(lr){4-5} \cmidrule(lr){6-7} \cmidrule(lr){8-9}
EWC++~\citep{ewc}           
& 57.34$\pm$2.10 & 60.33$\pm$2.73 
& 35.35$\pm$1.96 & 38.78$\pm$2.32 
& 22.26$\pm$1.15 & 24.39$\pm$1.18
& \rev{24.81} & \rev{26.21}\\
BiC~\citep{bic}             
& 58.38$\pm$0.54 & 61.49$\pm$0.68 
& 33.51$\pm$3.04 & 37.61$\pm$3.00 
& 22.80$\pm$0.94 & 24.90$\pm$1.07 
& \rev{27.41} & \rev{28.38} \\
ER-MIR~\citep{aljundi2019online}             
& 57.28$\pm$2.43 & 61.93$\pm$3.35 
& 35.35$\pm$1.41 & 38.28$\pm$1.15 
& 22.10$\pm$1.14 & 24.54$\pm$1.26 
& \rev{20.48} & \rev{20.68} \\ 
% A-GEM           
% & 39.29$\pm$2.88 & 44.85$\pm$4.70 & 33.12$\pm$5.03 
% & 4.62$\pm$0.23 & 6.94$\pm$0.51 & 15.11$\pm$1.03 
% & 9.42$\pm$2.15 & 12.14$\pm$3.58 & 16.63$\pm$1.39 \\
% \rev{GSS~\citep{aljundi2019gradient}}
% & \rev{55.51$\pm$3.33} & \rev{59.27$\pm$4.36} 
% & \rev{30.09$\pm$1.38} & \rev{35.06$\pm$1.43}
% & \rev{17.68$\pm$0.92} & \rev{20.79$\pm$0.99}
% & \rev{16.89} & \rev{19.01}
% \\
GDumb~\citep{Prabhu2020GDumbAS}           
& 53.20$\pm$1.93 & 55.27$\pm$2.69 
& 32.84$\pm$0.45 & 34.03$\pm$0.89 
& 18.17$\pm$0.19 & 18.69$\pm$0.45
& \rev{14.41} & \rev{14.21}\\
RM\textsuperscript{$\dagger$}~\citep{bang2021rainbow}
&23.00$\pm$1.43  & 61.52$\pm$3.69
& 8.63$\pm$0.19  &  33.27$\pm$1.59 
& 5.74$\pm$0.3  & 17.04$\pm$0.77 
& \rev{6.22} & \rev{28.30}\\ 
Baseline-ER (Sec.~\ref{sec:baseline})  
% Our Baseline (Sec.~\ref{sec:baseline})
& 57.46$\pm$2.25 & 60.17$\pm$2.96  
& 35.61$\pm$2.08 & 39.10$\pm$2.02 
& 22.45$\pm$1.15 & 24.54$\pm$1.26 
& \rev{25.16} & \rev{26.50} \\ 
\cmidrule(lr){1-1} \cmidrule(lr){2-3} \cmidrule(lr){4-5} \cmidrule(lr){6-7} \cmidrule(lr){8-9}    
\textbf{\method (Ours)}    
& \textbf{70.26$\pm$1.28} & \textbf{73.90$\pm$0.22} 
& \textbf{46.67$\pm$0.79} & \textbf{49.22$\pm$0.79} 
& \textbf{23.87$\pm$0.68} & \textbf{25.05$\pm$0.52} 
& \rev{\textbf{28.16}} & \rev{\textbf{28.88}}\\
\bottomrule
\\
\end{tabular}
}
\vspace{-1.5em}
\caption{Comparison of online CL methods on the i-Blurry setup for CIFAR10, CIFAR100, TinyImageNet \rev{and ImageNet}. \rev{`Joint Training\textsuperscript{$\ddagger$}' shows the final accuracy of non-CL joint training as a soft upper bound where all relevant hyper-parameters were kept consistent with other compared CL methods.} \method outperforms all other CL methods by large margins on both the $A_\text{AUC}$ and the $A_{avg}$.} % The best result for each of the metrics is shown in {\bf bold}.}
\vspace{-1em}
\label{tab:main}
\end{table}
\begin{figure}[t!]
    \centering
    \resizebox{1.0\linewidth}{!}{
    \includegraphics{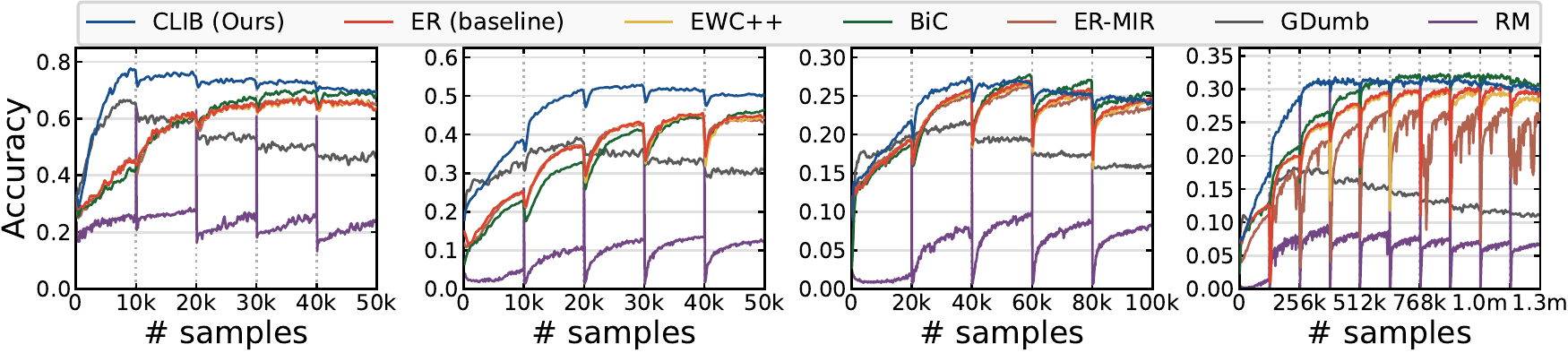}
    }
    {\footnotesize \hspace{0em}(a) CIFAR10 \hspace{4em}(b) CIFAR100 \hspace{4em}(c) TinyImageNet \hspace{4em}\rev{(d) ImageNet}}
    \vspace{-0.5em}
    \caption{\rev{Accuracy-to-\{number of samples\} for various CL methods on CIFAR10, CIFAR100, TinyImageNet and ImageNet} Our \method is consistent at maintaining high accuracy throughout inference while other CL methods are not as consistent.}
    \vspace{-2em}
    \label{fig:iblurry_plot}
\end{figure}
%-------------------------------------
\subsection{Results on the i-Blurry Setup}
In all our experiments, we denote the best result for each of the metrics in {\bf bold}.

We first compare various online CL methods in the i-Blurry setup on CIFAR10, CIFAR100, TinyImageNet \rev{and ImageNet} in Table~\ref{tab:main}.
On CIFAR10, proposed \method outperforms all other CL methods by large margins; at least $+12.67\%$ in $A_\text{AUC}$ and $+12.85\%$ in $A_{avg}$.
On CIFAR100, \method also outperforms all other methods by large margins; at least $+11.47\%$ in $A_\text{AUC}$ and $+10.18\%$ in $A_{avg}$.
On TinyImageNet, all methods score very low.
Nonetheless, \method outperforms other CL methods by at least $+2.71\%$ in $A_\text{AUC}$ and $+1.90\%$ in $A_{avg}$.
\rev{Surprisingly on ImageNet, most methods perform slightly better than on TinyImagenet. We believe the reason is that the samples per class is the same with TinyImageNet (\eg, 20) but ImageNet has higher resolution images, making learning from those images easier than TinyImageNet.
On ImageNet, \method still outperforms other CL methods by at least $+2.60\%$ in $A_\text{AUC}$ and $+2.06\%$ in $A_{avg}$.}
Since, \method uses memory only training scheme, the distribution of the training samples are stabilized through the memory (see Sec.~\ref{sec:memory_only_training}) which is helpful in the i-Blurry setup where the model encounters samples from more varied classes.
% In addition, the sample-wise importance based memory management helps to maintain samples from scarce classes, contributing to the performance of \method.

Note that Rainbow Memory (RM)~\citep{bang2021rainbow} exhibits a very different trend than other compared methods. 
On TinyImageNet, it performs poorly. 
We conjecture that the delayed learning from the two-stage training is particularly detrimental in larger datasets with longer training duration and tasks such as TinyImageNet. 
On CIFAR10 and CIFAR100, RM performs reasonably well in $A_{avg}$ but poorly in $A_\text{AUC}$.
This verifies that its two-stage training method delays most of the learning to the end of each task, resulting in a poor any-time inference performance measured by $A_\text{AUC}$.
% This training scheme is quite ineffective at any-time inference, as numerous inference queries that occur in the middle of the task will result in poor results.
Note that $A_{avg}$ fails to capture this; on CIFAR10, the difference in $A_{avg}$ for \method and RM is $+13.27\%$ which is similar to other methods but the difference for $A_\text{AUC}$ is $+47.55\%$ which is noticeably larger than other methods.
Similar trends can be found on \rev{other datasets} as well.

% RM highlights the importance of $A_\text{AUC}$ in accurately evaluating CL methods.
%
% This shows that the proposed method is a strong method for the i-Blurry setup on multiple datasets.
% Interestingly, A-GEM performs significantly worse than other methods in all the tested settings.
% This is because A-GEM is designed for the task-incremental setting where the task descriptors are available to the model.
% A-GEM utilizes the task descriptors to calculate gradient constraints which cannot be done effectively for our task-free setup.

We also show the accuracy-to-\{\# of samples\} curve for the CL methods on CIFAR10, CIFAR100, TinyImageNet \rev{and ImageNet} for comprehensive analysis throughout training in Fig~\ref{fig:iblurry_plot}.
Interestingly, RM shows a surged accuracy only at the task transitions due to its two-stage training method and the accuracy is overall low. %but spikes up at the end of the tasks due to the two-stage training method it uses.
Additionally, GDumb shows a \rev{severe} decreasing trend in accuracy as tasks progress.
It is because the `Dumb Learner' trains from scratch at every inference query leading to accuracy degradation.
In contrast, \method not only outperforms other methods but also shows the most consistent accuracy at all times.
\rev{More discussion of other methods are in Sec.~\ref{sec:comp_to_other_methods}.}%, as evidenced by the accuracy curve being above other methods for the majority of training.

\begin{table}[t!]
    \centering
    \resizebox{0.9\linewidth}{!}{
    \begin{tabular}{ccccccc}
    \toprule
    \multirow{2}{*}{Varying $N$} & \multicolumn{2}{c}{$N=0$ (Blurry)} & \multicolumn{2}{c}{$N=50$ (i-Blurry)}& \multicolumn{2}{c}{$N=100$ (Disjoint)}\\ 
    \cmidrule(lr){2-3} \cmidrule(lr){4-5} \cmidrule(lr){6-7}
         & $A_\text{AUC}$ & $A_\text{avg}$ & $A_\text{AUC}$ & $A_\text{avg}$ & $A_\text{AUC}$ & $A_\text{avg}$ \\ 
    \cmidrule(lr){1-1} \cmidrule(lr){2-3} \cmidrule(lr){4-5} \cmidrule(lr){6-7}
    EWC++    
    & 53.24$\pm$0.56 & 57.04$\pm$0.94 
    & 57.34$\pm$2.10 & 60.33$\pm$2.73  
    & 77.64$\pm$1.81 & 77.80$\pm$1.93  \\
    BiC      
    & 51.51$\pm$0.12 & 54.88$\pm$0.30 
    & 58.38$\pm$0.54 & 61.49$\pm$0.68 
    & {\bf 78.78$\pm$1.52} & {\bf 80.12$\pm$2.20} \\
    ER-MIR          
    & 52.21$\pm$0.85 & 56.33$\pm$0.47  
    & 57.28$\pm$2.43 & 61.93$\pm$3.35  
    & 76.49$\pm$1.97 & 78.20$\pm$2.01 \\ 
    % A-GEM           
    % & 41.25$\pm$0.58 & 49.07$\pm$0.97 & 21.50$\pm$8.37
    % & 39.29$\pm$2.88 & 44.85$\pm$4.70 & 33.12$\pm$5.03
    % & 44.97$\pm$2.27 & 44.80$\pm$1.36 & 94.95$\pm$1.77 \\
    % \rev{GSS}
    % & \rev{52.00$\pm$1.81} & \rev{54.87$\pm$1.21}
    % & \rev{55.51$\pm$3.33} & \rev{59.27$\pm$4.36}
    % & \rev{75.42$\pm$2.83} & \rev{75.36$\pm$3.08}\\
    GDumb           
    & 45.86$\pm$0.80 & 46.37$\pm$2.09 
    & 53.20$\pm$1.93 & 55.27$\pm$2.69 
    & 65.27$\pm$1.54 & 66.74$\pm$2.39 \\
    RM\textsuperscript{$\dagger$}
    &22.54$\pm$1.11  & 54.07$\pm$0.70
    &23.00$\pm$1.43  & 61.52$\pm$3.69
    &33.17$\pm$3.71  & 66.67$\pm$2.38 \\
    Baseline-ER   
    & 53.28$\pm$0.57 & 57.13$\pm$1.01 
    & 57.46$\pm$2.25 & 60.17$\pm$2.96
    & 77.82$\pm$2.06 & 77.47$\pm$2.69 \\
    \cmidrule(lr){1-1} \cmidrule(lr){2-3} \cmidrule(lr){4-5} \cmidrule(lr){6-7}
    \textbf{\method (Ours)}
    & {\bf 68.87$\pm$0.83} & {\bf72.79$\pm$0.96} 
    & \textbf{70.26$\pm$1.28} & \textbf{73.90$\pm$0.22}  
    &  78.58$\pm$2.09 & 77.96$\pm$3.28  \\
    \midrule
    \multirow{2}{*}{Varying $M$}& \multicolumn{2}{c}{$M=10$}& \multicolumn{2}{c}{$M=30$}&
    \multicolumn{2}{c}{$M=50$}
    \\ \cmidrule(lr){2-3} \cmidrule(lr){4-5} \cmidrule(lr){6-7}
         & $A_\text{AUC}$ & $A_\text{avg}$ & $A_\text{AUC}$ & $A_\text{avg}$ & $A_\text{AUC}$ & $A_\text{avg}$ \\ 
    \cmidrule(lr){1-1} \cmidrule(lr){2-3} \cmidrule(lr){4-5} \cmidrule(lr){6-7}
    EWC++    
    & 57.34$\pm$2.10 & 60.33$\pm$2.73
    & 65.71$\pm$2.20 & 69.94$\pm$3.01 
    & 68.01$\pm$0.85 & 73.26$\pm$2.33 \\
    BiC 
    & 58.38$\pm$0.54 & 61.49$\pm$0.68 
    & 65.88$\pm$3.24 & 70.31$\pm$4.88      
    & 68.08$\pm$3.72 & 73.33$\pm$1.21\\
    ER-MIR 
    & 57.28$\pm$2.43 & 61.93$\pm$3.35 
    & 65.99$\pm$2.28 & 70.47$\pm$3.41         
    & 68.13$\pm$0.65 & 73.33$\pm$1.21   \\ 
    % \rev{GSS}
    % & \rev{55.51$\pm$3.33} & \rev{59.27$\pm$4.36}
    % & \rev{65.19$\pm$2.65} & \rev{69.83$\pm$1.83}
    % & \rev{-} & \rev{-}\\
    % A-GEM           
    % & 44.97$\pm$2.27 & 44.80$\pm$1.36 & 94.95$\pm$1.77
    % & 39.29$\pm$2.88 & 44.85$\pm$4.70 & 33.12$\pm$5.03
    % & 50.13$\pm$5.40 & 56.12$\pm$5.54 & 29.59$\pm$12.94 \\
    GDumb            
    & 53.20$\pm$1.93 & 55.27$\pm$2.69 
    & 54.73$\pm$1.54 & 54.63$\pm$2.39
    & 53.86$\pm$0.59 & 52.82$\pm$1.24 \\
    RM\textsuperscript{$\dagger$}
    &23.00$\pm$1.43  & 61.52$\pm$3.69
    &26.60$\pm$1.74  & 61.52$\pm$0.48 
    &28.32$\pm$5.08  & 59.32$\pm$4.57 \\
    Baseline-ER   
    & 57.46$\pm$2.25 & 60.17$\pm$2.96
    & 65.92$\pm$2.25 & 70.04$\pm$2.87 
    & 68.26$\pm$0.84 & 73.16$\pm$1.49 \\
    \cmidrule(lr){1-1} \cmidrule(lr){2-3} \cmidrule(lr){4-5} \cmidrule(lr){6-7}
    \textbf{\method (Ours)}
    & \textbf{70.26$\pm$1.28} & \textbf{73.90$\pm$0.22}  
    & {\bf 75.04$\pm$2.81} & {\bf 77.87$\pm$2.57} 
    & {\bf 75.14$\pm$1.27} & {\bf 78.30$\pm$2.01} \\
    \bottomrule
    \end{tabular}
    }
    \vspace{-0.5em}
    \caption{Analysis on various values of $N$ (top) and $M$ (bottom) in the i-Blurry-$N$-$M$ setup using CIFAR10 dataset. For varying $N$, we use $M=10$. For varying $M$, we use $N=50$. Note that $N=0$ corresponds to the blurry split and $N=100$ corresponds to the disjoint split. For $N=100$, \method outperforms or performs on par with other CL methods. For $N=0$, the gap between \method and other methods widens. For $N=50$, \method again outperforms all comparisons by large margins.
    For varying $M$, \method outperforms all comparisons excepting only the $A_{avg}$ when $M=0$.}% The best result for each of the metrics is shown in {\bf bold}.}
    \vspace{-1.5em}
    \label{tab:class_dist}
\end{table}
%-----------------------------------------
\subsection{Analysis on Disjoint Class Percentages ($N$) and Blurry Levels ($M$).}
\label{sec:class_split_exp}
We further investigate the effect of different $N$ and $M$ values in the i-Blurry-N-M splits with various CL methods and summarize the results for varying values of the disjoint class percentages such as $N=0, 50, 100$ in Table~\ref{tab:class_dist} (top).
For $N=0$, \method outperforms other methods by at least $+16.59\%$ in $A_\text{AUC}$ and $+16.36\%$ in $A_{avg}$. 
For $N=100$, the performance is similar for the majority of the methods, with \method being the best in $A_\text{AUC}$.
For $N=50$, \method outperforms all comparisons by at least  $+12.55\%$ in $A_\text{AUC}$ and $+13.31\%$ in $A_{avg}$.
Even though \method was designed with the i-Blurry setup in mind, it also outperforms other CL methods in conventional setups such as the $N=100$ (disjoint) or the $N=0$ (blurry) setups.
It implies that \method is generally applicable to online CL setups and not restricted to just the i-Blurry setup.
Meanwhile, except GDumb, all methods show similar performance on the $N=100$ (disjoint) setup. 
The results imply that the i-Blurry setup differentiates CL methods better than the more traditionally used disjoint setup.

Following~\cite{bang2021rainbow}, we additionally summarize the results for varying values of the blurry level such as $M=10, 30, 50$ in Table~\ref{tab:class_dist} (bottom).
We observe that \method again outperforms or performs on par with other CL methods on various blurry levels. 

%----------------------------------------
\subsection{Ablation Studies}
\label{sec:ablations}
We show the ablation studies (on CIFAR10/100) for each of the proposed components in Table~\ref{tab:ablation}.

\begin{table}[t!]
    \centering
    \resizebox{1.0\linewidth}{!}{
\begin{tabular}{lcccc}
\toprule
\multirow{2}{*}{Methods} & \multicolumn{2}{c}{CIFAR10} & \multicolumn{2}{c}{CIFAR100}\\ \cmidrule(lr){2-3} \cmidrule(lr){4-5}
     & $A_\text{AUC}$ & $A_{avg}$ & $A_\text{AUC}$ & $A_{avg}$ \\ 
\midrule
\method 
& \textbf{70.26$\pm$1.28} & \textbf{73.90$\pm$0.22}
& \textbf{46.67$\pm$0.79} & \textbf{49.22$\pm$0.79}   \\    
~~w/o Sample Importance Mem. (Sec.\ref{sec:per_sample})
& 53.75$\pm$2.11 & 56.31$\pm$2.56 
& 36.59$\pm$1.22 & 38.59$\pm$1.21    \\
~~w/o Memory-only training (Sec.\ref{sec:memory_only_training})
& 67.06$\pm$1.51 & 71.65$\pm$1.87 
& 44.63$\pm$0.22 & 48.66$\pm$0.39  \\
~~w/o Adaptive LR scheduling (Sec.\ref{sec:adaptive_lr})
& 69.70$\pm$1.34 & 73.06$\pm$1.35 
& 45.01$\pm$0.22 & \rev{48.97$\pm$0.12}  \\
%\method (Ours) & \textbf{70.20} & \textbf{74.46} &   & \textbf{46.78} & \textbf{49.51} &   \\
\bottomrule
\\
\end{tabular}
}
\vspace{-2em}
\caption{Ablations for proposed components of our method using CIFAR10 and CIFAR100 dataset. All proposed components improve the performance, with sample-wise importance based memory management providing the biggest gains. While adaptive LR scheduling provides small gains in CIFAR10, the gains increase in the more challenging CIFAR100.} % The best result for each of the metrics is shown in {\bf bold}.}
\vspace{-0.5em}
\label{tab:ablation}
\end{table}

\textbf{Sample-wise Importance Based Memory Management.}
We replace the `sample-wise importance memory management' module with the reservoir sampling.
As shown in the table, the removal of our memory management strategy degrades the performance in both $A_\text{AUC}$ and $A_{avg}$ on both CIFAR10 and CIFAR100.
As explained in Sec.~\ref{sec:per_sample}, reservoir sampling removes samples at random, hence samples are discarded without considering if some samples are more important than others.
Thus, using sample-wise importance to select which sample to discard greatly contributes to performance.

\textbf{Memory Only Training.}
We replace the memory usage strategy from our memory only training with ER.
Training with ER means that samples from both the online stream and the memory are used.
Without the proposed memory only training scheme, the performance degrades across the board by fair margins. %, although not as much as the per sample memory management.
As the streamed samples are being used directly without the sample memory acting as a distribution regularizer (see Sec.~\ref{sec:memory_only_training}), the CL model is more influenced by the recently observed samples, skewing the training and resulting in worse performance.

\textbf{Adaptive Learning Rate Scheduling.}
We change the LR scheduling from adaptive LR to exponential with reset.
The performance drop is larger in the more challenging CIFAR100 where more training iterations make the ablated model suffer more from a lack of good adaptive LR scheduling.

%------------------------
\section{Conclusion}
\label{sec:conclusion}
We question the practicality of existing continual learning setups for real-world application and propose a novel CL setup named i-Blurry. 
It is online, task-free, class-incremental, has blurry task boundaries, and is subject to any-time inference.
Additionally, we propose a new metric to better evaluate the effectiveness of any-time inference.
To address this realistic CL setup, we propose a method which uses per-sample memory management, memory only training, and adaptive LR scheduling, named \methodfull (\method).
Our proposed \method consistently outperforms existing CL methods in multiple datasets and setting combinations by large margins.

\section*{Ethics Statement}

All continual learning (CL) methods including the proposed one would adapt and extend the already trained AI model to recognize better with the streamed data.
The CL methods will expedite the deployment of AI systems to help humans by its versatility of adapting to a new environment out of the factory or research labs.
As all CL methods, however, would suffer from adversarial streamed data as well as data bias, which may cause ethnic, gender or biased gender issues, the proposed \method would not be an exception.
Although the proposed \method has \emph{no intention} to allow such problematic cases, the method may be exposed to such threats.
Relentless efforts should be made to develop mechanisms to prevent such usage cases in order to make the continuously updating machine learning models safer and enjoyable to be used by humans.

\section*{Reproducibility Statement}

We take the reproducibility of the research very seriously and release all codes, data splits and containers (\eg, Docker) that include the general framework, learned models, and downstream tasks at \url{https://github.com/naver-ai/i-Blurry}.

\section*{Acknowledgement}
This work was partly supported by the National Research Foundation of Korea (NRF) grant funded by the Korea government (MSIT) (No.2022R1A2C4002300) and Institute for Information \& communications Technology Promotion (IITP) grants funded by the Korea government (MSIT) (No.2020-0-01361-003 and 2019-0-01842, Artificial Intelligence Graduate School Program (Yonsei University, GIST), and No.2021-0-02068 Artificial Intelligence Innovation Hub)). 

%% file: appendix_camready.tex
\rev{\subsection{Additional Discussion for our Baseline}
\label{sec:baseline}
In constructing our baseline, we used the reservoir sampling for memory management and ER for memory usage.
We used reservoir sampling as it is widely used in the online and task-free setups with good performance.
Note that in online CL, memory management policies that use the entire task's samples at once, such as herding selection~\citep{icarl}, mnemonics~\citep{mnemonics}, and rainbow memory~\citep{bang2021rainbow} are inapplicable.\\
For the memory usage, we use ER which draws half of the training batch from the stream and the other half from the memory, following a large number of online CL methods based on ER with good performance~\citep{mai2021online}. \\
For the LR scheduling, we first note that other CL methods use either \rev{(1) exponential decay}~\citep{icarl, ewc, mirzadeh2020understanding} or (2) constant LR. 
We do not use (1) as it is hyper-parameter sensitive; the decay rate that works for CIFAR10 decayed the LR too quickly for larger datasets such as CIFAR100.
If the LR is decayed too fast, the LR becomes too small to learn about new classes that are introduced in the future.
Thus, we use exponential LR scheduler with the modification that the LR is reset when a new class is observed.
Comparing with the constant LR, we obtain slightly better performance for EWC++ and our baseline on CIFAR10, as shown in Table \ref{tab:exp_reset}. Thus, we denote this LR schedule as the exponential with reset and use it in our baseline.}
\subsection{Proof of Theorem~\ref{th:per_sample}}
\label{sec:proof_of_th1}
We give the proof of Theorem~\ref{th:per_sample} below.
Our assumption is that when selecting memory $\mathcal{M}$ from a set of candidates $\mathcal{C}$, we should select $M$ so that optimizing on $M$ maximizes the loss decrease on $C$. In equations, optimal memory $\mathcal{M}^*$ is
\begin{align}
    \mathcal{M}^* &= \argmax_{\mathcal{M}\subset\mathcal{C},\, |\mathcal{M}|\leq m}  \sum_{(x', y')\in\mathcal{M}}\sum_{(x, y)\in\mathcal{C}}\mathbb{E}_\theta\left[l(x, y;\,\theta) - l(x, y;\,\theta-\nabla_\theta l(x', y';\,\theta))\right]\\
    &= \argmax_{\mathcal{M}\subset\mathcal{C},\, |\mathcal{M}|\leq m}\left[ \sum_{(x', y')\in\mathcal{C}}\sum_{(x, y)\in\mathcal{C}}\mathbb{E}_\theta\left[l(x, y;\,\theta) - l\left(x, y;\, \theta-\nabla_\theta l(x', y';\,\theta)\right)\right]\right.\nonumber\\
    &\qquad\qquad\qquad - \left.\sum_{(x', y')\in\mathcal{C}\setminus\mathcal{M}}\sum_{(x, y)\in\mathcal{C}}\mathbb{E}_\theta\left[l(x, y;\,\theta) - l\left(x, y;\,\theta-\nabla_\theta l(x', y';\,\theta)\right)\right]\right]\\
    &= \argmin_{{\mathcal{M}\subset\mathcal{C},\, |\mathcal{M}|\leq m}}\sum_{(x', y')\in\mathcal{C}\setminus\mathcal{M}}\sum_{(x, y)\in\mathcal{C}}\mathbb{E}_\theta\left[l(x, y;\,\theta) - l\left(x, y;\,\theta-\nabla_\theta l(x', y';\,\theta)\right)\right],
\end{align}
where $\theta$ is the model parameter, $l$ is the loss function, and $m$ is the memory size.
Since we perform memory update after every streamed sample, the problem reduces to selecting one sample $(\bar{x}, \bar{y})$ to remove when the memory is full. Thus, $\mathcal{C}\setminus\mathcal{M} = \{(\bar{x}, \bar{y})\}$.
The optimal removed sample $(\bar{x}^*, \bar{y}^*)$ would be
\begin{equation}
\label{eq:optimal_remove}
    (\bar{x}^*, \bar{y}^*) = \argmin_{(\bar{x}, \bar{y})\in\mathcal{C}} \mathbb{E}_\theta\left[\sum_{(x, y)\in\mathcal{C}}l(x, y;\, \theta) - l\left(x, y;\, \theta-\nabla_\theta l(\bar{x}, \bar{y};\, \theta)\right)\right].
\end{equation}

\subsection{Details on Sample-wise Importance Based Memory Management}
\label{sec:psi_mem_app}
We describe the details of our sample-wise importance based memory management here. We update $\mathcal{H}$, the estimate of sample-wise importance for episodic memory, after every model update.
The details are in Alg.~\ref{algo:loss_history}.
\rev{The runtime complexity of updating the sample-wise importance is $\mathcal{O}(M)$ where $M$ is the fixed memory size. Because the memory size is kept constant regardless of the data size, the algorithm can scale to larger datasets. Note that for practical efficiency, Alg. 1 can be used after every $k$ model updates instead of every model update as described above.} 
With the sample-wise importance scores, we update the memory everytime a new sample is encountered.
The details are in Alg.~\ref{algo:memory_update}.

\begin{algorithm*}[t!]
\caption{Update Sample-wise Importance}
\label{algo:loss_history}
\begin{algorithmic}[1]
\State \textbf{Input} model $f_\theta$, memory $\mathcal{M}$, sample-wise importance $\mathcal{H}$, previous loss $l_\text{prev}$, indices used for training $\mathcal{I}$, update coefficient $\lambda$
\State $l_\text{cur} = \frac{1}{|\mathcal{M}|}\sum_{(x, y)\in \mathcal{M}}l(x, y;\,\theta)$ {\small\color{azure}\Comment{Obtain memory loss}}
\State $\Delta l= l_\text{prev} - l_\text{cur}${\small\color{azure}\Comment{Obtain memory loss decrease}}
\State $\Delta l_\text{pred} = \frac{1}{|\mathcal{I}|}\sum_{i \in \mathcal{I}}\mathcal{H}_{i}${\small\color{azure}\Comment{Memory loss decrease prediction using current $\mathcal{H}$}}
\For{$i \in \mathcal{I}$}
    \State \textbf{Update} $\mathcal{H}_{i} \leftarrow \mathcal{H}_{i} + \lambda(\Delta l-\Delta l_\text{pred})${\small\color{azure}\Comment{Update $\mathcal{H}$ for samples used for training}}
\EndFor
\State \textbf{Update} $l_\text{prev} \leftarrow l_\text{cur}$
\State \textbf{Output} $\mathcal{H}$, $l_\text{prev}$
\end{algorithmic}
\end{algorithm*}

\begin{algorithm*}[t!]
\caption{Sample-wise Importance Based Memory Update}
\label{algo:memory_update}
\begin{algorithmic}[1]
\State \textbf{Input} model $f_\theta$, memory $\mathcal{M}$, memory size $m$, sample $(\hat{x}, \hat{y})$, per-sample criterion $\mathcal{H}$, previous loss $l_\text{prev}$
\If{$|\mathcal{M}|<m$} {\small\color{azure}\Comment{If the memory is not full}}
    \State \textbf{Update} $\mathcal{M} \leftarrow\mathcal{M}\cup\{(\hat{x}, \hat{y})\}$ {\small\color{azure}\Comment{Append the sample to the memory}}
    \State $\bar{i} = |\mathcal{M}|$
\Else {\small\color{azure}\Comment{If the memory is already full}}
    \State $y_\text{max} = \argmax_y|\{(x_i, y_i)|(x_i, y_i)\in{\mathcal{M}\cup\{(\hat{x}, \hat{y})\}},\, y_i=y\}|$ {\small\color{azure}\Comment{Find the most frequent label}}
    \State $\mathcal{I}_{y_\text{max}}=\{i|(x_i, y_i)\in\mathcal{M},\, y_i=y_\text{max}\}$
    \State $\hat{i} = \argmin_{i \in \mathcal{I}_{y_\text{max}}} \mathcal{H}_i$ {\small\color{azure}\Comment{Find the sample with the lowest importance}}
    \State \textbf{Update} $l_\text{prev} \leftarrow \frac{m}{m-1}l_\text{prev} - \frac{1}{m-1}l(x_{\hat{i}}, y_{\hat{i}};\,\theta)$
    \State \textbf{Update} $\mathcal{M}_{\hat{i}} \leftarrow (\hat{x}, \hat{y})${\small\color{azure}\Comment{Replace that sample with the new sample}}
\EndIf
\State \textbf{Update} $l_\text{prev} \leftarrow \frac{|\mathcal{M}|-1}{|\mathcal{M}|}l_\text{prev} + \frac{1}{|\mathcal{M}|}l\left(\theta, \hat{x}, \hat{y}\right)$
\State $\mathcal{I}_{\hat{y}}=\{i|(x_i, y_i)\in \mathcal{M},  y_i=\hat{y}, i \neq \hat{i}\}$
\State \textbf{Update} $\mathcal{H}_{\hat{i}} \leftarrow \frac{1}{|\mathcal{I}_{\hat{y}}|}\sum_{i \in \mathcal{I}_{\hat{y}}}\mathcal{H}_i$ {\small\color{azure}\Comment{Initialize the importance for the new sample}}
\State \textbf{Output} $\mathcal{M}$, $\mathcal{H}$, $l_\text{prev}$

\end{algorithmic}
\end{algorithm*}

\subsection{Adaptive Learning Rate Scheduler}
\label{sec:adaptive_lr_alg}
We describe the adaptive LR schedule from \ref{sec:adaptive_lr} in Alg.~\ref{sec:adaptive_lr_alg}. We fix the significance level to the commonly used $\alpha=0.05$.
Our adaptive LR scheduling decreases or increases the LR based on its current value.
Thus, the rate in which the LR can change is bounded and sudden changes in LR do not happen.
\begin{algorithm*}[t!]
\caption{Adaptive Learning Rate Scheduler}
\label{algo:lr_schedule}
\begin{algorithmic}[1]
\State \textbf{Input} current LR $\eta$, current base LR $\bar{\eta}$, loss before applying current LR $l_\text{before}$, current loss $l_\text{cur}$, LR performance history $\mathcal{H}^\text{(high)}$ and $\mathcal{H}^\text{(low)}$, LR step $\gamma<1$, history length $m$, significance level $\alpha$

\State $l_\text{diff} = l_\text{before} - l_\text{cur}$ {\small\color{azure}\Comment{Obtain loss decrease}}
\State \textbf{Update} $l_\text{before} \leftarrow l_\text{cur}$
\If{$\eta>\bar{\eta}$}{\small\color{azure}\Comment{If LR is higher than base LR}}
    \State \textbf{Update} $\mathcal{H}^\text{(high)} \leftarrow \mathcal{H}^\text{(high)}\cup\{l_\text{diff}\}${\small\color{azure}\Comment{Append loss decrease in high LR history}}
    \If{$|\mathcal{H}^\text{(high)}| > m$}
        \State \textbf{Update} $\mathcal{H}^\text{(high)} \leftarrow \mathcal{H}^\text{(high)}\setminus\{\mathcal{H}^\text{(high)}_1\}$
    \EndIf
\Else{\small\color{azure}\Comment{If LR is lower than base LR}}
    \State \textbf{Update} $\mathcal{H}^\text{(low)} \leftarrow \mathcal{H}^\text{(low)}\cup\{l_\text{diff}\}$ {\small\color{azure}\Comment{Append loss decrease in low LR history}}
    \If{$|\mathcal{H}^\text{(low)}| > m$}
        \State \textbf{Update} $\mathcal{H}^\text{(low)} \leftarrow \mathcal{H}^\text{(low)}\setminus\{\mathcal{H}^\text{(low)}_1\}$ 
    \EndIf
\EndIf
\If{$|\mathcal{H}^\text{(high)}| = m \textbf{ and } |\mathcal{H}^\text{(low)}| = m$}{\small\color{azure}\Comment{If both histories are full}}
    \State $p = \text{OneSidedStudentsTTest}\left(\mathcal{H}^\text{(low)}, \mathcal{H}^\text{(high)} \right)${\small\color{azure}
    \Statex\Comment{Perform one-sided Student's $t$-test with alternative hypothesis $\mu_\text{low} > \mu_\text{high}$}}
    \If{$p < \alpha$}  {\small\color{azure}\Comment{If pvalue is significantly low}}
        \State \textbf{Update} $\bar{\eta} \leftarrow \gamma^2\cdot\bar{\eta}${\small\color{azure}\Comment{Decrease base LR}}
        \State \textbf{Update} $\mathcal{H}^\text{(low)}, \mathcal{H}^\text{(high)} = \emptyset, \emptyset${\small\color{azure}\Comment{Reset histories}}
    \ElsIf{$p > 1-\alpha$}{\small\color{azure}\Comment{If pvalue is significantly high}}
        \State \textbf{Update} $\bar{\eta} \leftarrow \frac{1}{\gamma^2}\cdot\bar{\eta}${\small\color{azure}\Comment{Increase base LR}}
        \State \textbf{Update} $\mathcal{H}^\text{(low)}, \mathcal{H}^\text{(high)} = \emptyset, \emptyset$
    \EndIf
\EndIf
\If{$\eta > \bar{\eta}$}{\small\color{azure}\Comment{Alternately apply high and low LR (note that $\gamma<1$)}}
    \State \textbf{Update} $\eta = \gamma \cdot \bar{\eta}$
\Else
    \State \textbf{Update} $\eta = \frac{1}{\gamma} \cdot \bar{\eta}$
\EndIf
\State \textbf{Output} $\eta, \bar{\eta}, l_\text{before}, \mathcal{H}^\text{(low)}, \mathcal{H}^\text{(high)}$
\end{algorithmic}
\end{algorithm*}

\subsection{Details on the Online Versions of Compared CL Methods}
\label{sec:app_online_detail}
We implemented online versions of RM~\citep{bang2021rainbow}, EWC++~\citep{rwalk}, BiC~\citep{bic}, GDumb~\citep{Prabhu2020GDumbAS}, A-GEM~\citep{AGEM}, GSS~\citep{aljundi2019gradient} and MIR~\citep{aljundi2019online} by incorporating ER and exponential decay with reset to the methods whenever possible. There is no specified memory management strategy for EWC++, and BiC uses herding selection from iCaRL\citep{icarl}. However, herding selection is not possible in online since it requires whole task data for calculating class mean, so we attach reservoir memory to both methods instead. EWC++ does not require any other modification.

Additional modification should be applied to bias correction stage of BiC. BiC originally performs bias correction at end of each task, but since evaluation is also performed at the middle of task in our setup, we modified the method to perform bias correction whenever the model receives inference query.

In RM, their memory management strategy based on uncertainty is not applicable in an online setup, since it requires uncertainty rankings of the whole task samples. Thus, we replace their sampling strategy with balanced random sampling, while keeping their two-stage training scheme. Methods that were converted from offline to online, namely EWC++, BiC, and RM, may have suffered some performance drop due to deviation from their original methods.

\subsection{Analysis on Sample Memory Size \rev{and Number of Tasks}}
\label{sec:app_memory_size}
%-----------------------------------------

We conduct analysis over various sample memory sizes ($K$) and summarize results in Table~\ref{tab:memory_size}.
We observe that \method outperforms other CL methods in both $A_\text{AUC}$ and $A_{avg}$ no matter the memory size.
It is interesting to note that \method with a memory size of $K=200$ outperforms other CL methods with a memory size of $K=1000$ in the $A_\text{AUC}$ and performs on par in $A_{avg}$.
Thus, \method is the only method using memory only training scheme but is the least sensitive to memory size. 
It implies that our memory management policy is the most effective, which shows the superiority of our per-sample memory management method.

\begin{table}[t!]
    \centering
    \resizebox{1.0\linewidth}{!}{
\begin{tabular}{ccccccc}
\toprule
\multirow{2}{*}{Methods} & \multicolumn{2}{c}{K=200} & \multicolumn{2}{c}{K=500}            & \multicolumn{2}{c}{K=1000}\\ 
\cmidrule(lr){2-3} \cmidrule(lr){4-5} \cmidrule(lr){6-7}
     & $A_\text{AUC}$ & $A_\text{avg}$ & $A_\text{AUC}$ & $A_\text{avg}$ & $A_\text{AUC}$ & $A_\text{avg}$\\ 
\cmidrule(lr){1-1} \cmidrule(lr){2-3} \cmidrule(lr){4-5} \cmidrule(lr){6-7}
EWC++           
& 52.06$\pm$2.24 & 54.09$\pm$3.57 
& 57.34$\pm$2.10 & 60.33$\pm$2.73 
& 60.93$\pm$1.02 & 65.86$\pm$2.05  \\
BiC             
& 53.00$\pm$1.03 & 54.36$\pm$1.64
& 58.38$\pm$0.54 & 61.49$\pm$0.68 
& 61.52$\pm$2.24 & 64.82$\pm$1.15  \\
ER-MIR          
& 51.63$\pm$2.43 & 54.40$\pm$3.50 
& 57.28$\pm$2.43 & 61.93$\pm$3.35 
& 61.18$\pm$1.08 & 66.05$\pm$2.29 \\ 
% \rev{GSS}
% & \rev{49.53$\pm$1.10} & \rev{51.99$\pm$1.61}
% & \rev{55.51$\pm$3.33} & \rev{59.27$\pm$4.36}
% & \rev{60.60$\pm$1.00} & \rev{64.23$\pm$2.21} \\
% A-GEM           
% & 37.38$\pm$2.55 & 43.15$\pm$5.30 & 35.28$\pm$7.26 
% & 39.29$\pm$2.88 & 44.85$\pm$4.70 & 33.12$\pm$5.03
% & 40.20$\pm$3.67 & 45.90$\pm$6.11 & 32.50$\pm$8.70 \\
GDumb           
& 42.54$\pm$2.01 & 43.99$\pm$2.28 
& 53.20$\pm$1.93 & 55.27$\pm$2.69 
& 66.55$\pm$1.10 & 69.21$\pm$1.29 \\
RM\textsuperscript{$\dagger$}
&21.24$\pm$1.35  & 46.79$\pm$3.78
&23.00$\pm$1.43  & 61.52$\pm$3.69
& 26.13$\pm$1.61  & 72.29$\pm$2.17 \\
Baseline-ER   
& 52.11$\pm$2.32 & 54.34$\pm$3.34 
& 57.46$\pm$2.25 & 60.17$\pm$2.96  
& 61.18$\pm$1.08 & 66.05$\pm$2.29  \\ 
\cmidrule(lr){1-1} \cmidrule(lr){2-3} \cmidrule(lr){4-5} \cmidrule(lr){6-7}
\textbf{\method (Ours)}
& {\bf64.67$\pm$1.86} & {\bf 66.06$\pm$1.78} 
& \textbf{70.26$\pm$1.28} & \textbf{73.90$\pm$0.22} 
& {\bf73.00$\pm$1.30} & {\bf77.87$\pm$1.15}  \\
\bottomrule
\\
\end{tabular}
}
\vspace{-2em}
\caption{Analysis on various sample memory sizes ($K$) using CIFAR10. The i-Blurry-50-10 splits are used. The results are averaged over 3 runs. \method outperforms all other CL methods by large margins for all the memory sizes. \method uses the given memory budget most effectively, showing the superiority of our per-sample memory management method.} % The best result for each of the metrics is shown in {\bf bold}.}
\label{tab:memory_size}
\end{table}

\rev{
\begin{table}[h!]
    \centering
    \resizebox{1.0\linewidth}{!}{
\begin{tabular}{ccccccc}
\toprule
\multirow{2}{*}{Methods} & \multicolumn{2}{c}{5 Tasks} & \multicolumn{2}{c}{10 Tasks}  & \multicolumn{2}{c}{25 Tasks}\\ \cmidrule(lr){2-3} \cmidrule(lr){4-5} \cmidrule(lr){6-7}
     & $A_\text{AUC}$ & $A_\text{avg}$ & $A_\text{AUC}$ & $A_\text{avg}$ & $A_\text{AUC}$ & $A_\text{avg}$ \\
\cmidrule(lr){1-1} \cmidrule(lr){2-3} \cmidrule(lr){4-5} \cmidrule(lr){6-7}    
EWC++          
& 35.35$\pm$1.96 & 38.78$\pm$2.32 
& 32.25$\pm$1.56 & 34.85$\pm$1.71
& 27.08$\pm$1.56 & 29.46$\pm$1.65  \\
BiC            
& 33.51$\pm$3.04 & 37.61$\pm$3.00 
& 35.29$\pm$1.07 & 37.40$\pm$1.04 
& 34.68$\pm$1.16 & 35.55$\pm$1.38  \\
ER-MIR       
& 35.35$\pm$1.41 & 38.28$\pm$1.15 
& 33.65$\pm$1.51 & 35.29$\pm$1.56 
& 28.16$\pm$1.29 & 30.00$\pm$1.17  \\
% GSS
% & 30.09$\pm$1.38 & 35.06$\pm$1.43
% & 21.56$\pm$0.83 & 25.61$\pm$0.92 
% & 14.53$\pm$0.54 & 17.85$\pm$0.57  \\
GDumb
& 32.84$\pm$0.45 & 34.03$\pm$0.89         
& 31.22$\pm$0.51 & 32.35$\pm$1.73 
& 30.79$\pm$0.90 & 31.87$\pm$1.08  \\
RM\textsuperscript{$\dagger$}
& 8.63$\pm$0.19  &  33.27$\pm$1.59 
& 6.56$\pm$0.43  & 34.93$\pm$3.94
& 3.66$\pm$0.10  & 36.85$\pm$1.17 \\ 
Baseline-ER
& 35.61$\pm$2.08 & 39.10$\pm$2.02 
& 32.46$\pm$1.35 & 34.76$\pm$1.34 
& 28.35$\pm$1.56 & 30.45$\pm$1.43  \\ 
\cmidrule(lr){1-1} \cmidrule(lr){2-3} \cmidrule(lr){4-5} \cmidrule(lr){6-7}     
\textbf{\method (Ours)}    
& \textbf{46.67$\pm$0.79} & \textbf{49.22$\pm$0.79} 
& \textbf{44.61$\pm$1.16} & \textbf{46.15$\pm$1.07} 
& \textbf{43.09$\pm$1.04} & \textbf{43.64$\pm$1.08}  \\
\bottomrule
\\
\end{tabular}
}
\vspace{-2em}
\caption{\rev{Analysis on the number of tasks on CIFAR100. The i-Blurry-50-10 splits are used. The results are averaged over 3 runs. \method outperforms all other CL methods by large margins, even when the task sequences becomes longer. Additionally, \method does not suffer severe performance drops as the number of tasks increase, indicating that \method is well-suited for long-run CL problem setups as well.}}
\vspace{-1em}
\label{tab:numtasks}
\end{table}
}

\rev{
We additionally conduct analysis on the number of tasks used for the CIFAR100 dataset and summarize the results in Table~\ref{tab:numtasks}.
Note that we use CIFAR100 as CIFAR10 has too few classes to divide into longer task sequences.
We observe that \method outperforms other CL methods in both $A_\text{AUC}$ and $A_{avg}$ even when the number of tasks increase from 5 to 10 or 25.
Moreover, \method shows only minor performance drops with increasing number of tasks, indicating that \method is capable of achieving high performance in the long-run.\\\\
Interestingly, we note that BiC shows a slight performance increase when the number of tasks is increased to 10 or 25.
We believe this is because BiC uses separate bias parameters for each task, longer task sequences allows more fine-grained bias correction and may actually be favorable for BiC.
We also point out that the absolute performance of BiC is still lacking. % as it is often the worst or the second worst performing CL method.
Thus, it may imply that the bias correction of BiC contributes to strong stability in the stability-plasticity trade-off~\citep{Chaudhry_2018_ECCV}, which help maintain performance for longer task sequences where forgetting would be more severe.
However, as stability is enforced strongly, BiC might lack plasticity to sufficiently learn from new data, resulting in low performance overall.\\\\
We also note that EWC++ and Baseline-ER have the same $A_{AUC}$ and $A_{avg}$ values up to the second decimal places in the 10 tasks setting. 
This may imply that EWC++ is seldom beneficial over Baseline-ER as the added episodic memory negates the need for the regularization in EWC++.}

\subsection{Additional Comparisons with A-GEM}
\label{sec:app_a_gem}
We present additional comparisons to A-GEM.
Note that as A-GEM was designed for the task-incremental setting, it performs very poorly in our i-Blurry setup which is task-free.
Notably, it achieves only $4.62$ $A_\text{AUC}$ and $6.94$ $A_{avg}$ on CIFAR100, but other works~\citep{Prabhu2020GDumbAS,mai2021online} have also reported very poor performance for A-GEM in their studies as well.

\begin{table}[t!]
    \centering
    \resizebox{0.8\linewidth}{!}{
\begin{tabular}{ccccc}
\toprule
\multirow{2}{*}{Methods} & \multicolumn{2}{c}{CIFAR10} & \multicolumn{2}{c}{CIFAR100}\\
\cmidrule(lr){2-3} \cmidrule(lr){4-5}
     & $A_\text{AUC}$ & $A_\text{avg}$ & $A_\text{AUC}$ & $A_\text{avg}$ \\ \midrule
EWC++           
& 57.34$\pm$2.10 & 60.33$\pm$2.73 
& 35.35$\pm$1.96 & 38.78$\pm$2.32  \\
BiC             
& 58.38$\pm$0.54 & 61.49$\pm$0.68 
& 33.51$\pm$3.04 & 37.61$\pm$3.00   \\
ER-MIR             
& 57.28$\pm$2.43 & 61.93$\pm$3.35 
& 35.35$\pm$1.41 & 38.28$\pm$1.15  \\ 
A-GEM           
& 39.29$\pm$2.88 & 44.85$\pm$4.70 
& 4.62$\pm$0.23 & 6.94$\pm$0.51 \\
% \rev{GSS}
% & \rev{55.51$\pm$3.33} & \rev{59.27$\pm$4.36} 
% & \rev{30.09$\pm$1.38} & \rev{35.06$\pm$1.43} \\
GDumb           
& 53.20$\pm$1.93 & 55.27$\pm$2.69 
& 32.84$\pm$0.45 & 34.03$\pm$0.89  \\
RM\textsuperscript{$\dagger$}
&23.00$\pm$1.43  & 61.52$\pm$3.69
& 8.63$\pm$0.19  &  33.27$\pm$1.59  \\ 
Baseline-ER   
& 57.46$\pm$2.25 & 60.17$\pm$2.96  
& 35.61$\pm$2.08 & 39.10$\pm$2.02 \\
\cmidrule(lr){1-1} \cmidrule(lr){2-3} \cmidrule(lr){4-5}
\textbf{\method (Ours)}    
& \textbf{70.26$\pm$1.28} & \textbf{73.90$\pm$0.22} 
& \textbf{46.67$\pm$0.79} & \textbf{49.22$\pm$0.79}  \\
\bottomrule
\\
\end{tabular}
}
\caption{Additional comparisons to A-GEM with various online CL methods on the i-Blurry setup for CIFAR10 and CIFAR100 are shown. The i-Blurry-50-10 splits are used for all the datasets and the results are averaged over 3 runs. A-GEM performs very poorly, especially on CIFAR100 as it was designed for the task-incremental setting whereas i-Blurry setup is task-free. \method outperforms all other CL methods by large margins on both the $A_\text{AUC}$ and the $A_{avg}$.} % The best result for each of the metrics is shown in {\bf bold}.}
\label{tab:main_agem}
\end{table}

\subsection{Comparisons to Other Memory Management CL Methods}
\label{sec:app_gss}
We present additional comparison to CL methods that use a different memory management strategy in Table~\ref{tab:app_gss}.
GSS~\citep{aljundi2019gradient} is added as an additional comparison while Baseline-ER is used to represent the reservoir sampling.
\method outperforms both methods by large margins in both $A_\text{AUC}$ and $A_{avg}$, implying that the sample-wise importance memory management method is better than reservoir or GSS-greedy.

\begin{table}[t!]
    \centering
    \resizebox{1.0\linewidth}{!}{
\begin{tabular}{cccccc}
\toprule
\multirow{2}{*}{Methods}& \multirow{2}{*}{Mem. Management} & \multicolumn{2}{c}{CIFAR10} & \multicolumn{2}{c}{CIFAR100}\\
\cmidrule(lr){3-4} \cmidrule(lr){5-6}
&     & $A_\text{AUC}$ & $A_\text{avg}$ & $A_\text{AUC}$ & $A_\text{avg}$ \\ \midrule
% EWC++           
% & 57.06$\pm$1.63 & 59.96$\pm$3.18 
% & 35.26$\pm$1.82 & 39.35$\pm$1.50  \\
% BiC             
% & 52.35$\pm$1.85 & 54.94$\pm$1.81 
% & 26.26$\pm$0.92 & 29.90$\pm$1.03  \\
% ER-MIR             
% & 57.65$\pm$2.49 & 61.15$\pm$3.74 
% & 34.49$\pm$2.29 & 37.95$\pm$2.05  \\ 
% A-GEM           
% & 39.29$\pm$2.88 & 44.85$\pm$4.70 
% & 4.62$\pm$0.23 & 6.94$\pm$0.51 \\
GSS & GSS-Greedy
& 55.51$\pm$3.33 & 59.27$\pm$4.36 
& 30.09$\pm$1.38 & 35.06$\pm$1.43 \\
% GDumb           
% & 53.20$\pm$1.93 & 55.27$\pm$2.69 
% & 32.84$\pm$0.45 & 34.03$\pm$0.89 \\
Baseline-ER   & Reservoir 
& 57.46$\pm$2.25 & 60.17$\pm$2.96  
& 35.61$\pm$2.08 & 39.10$\pm$2.02   \\
\cmidrule(lr){1-1} \cmidrule(lr){2-2} \cmidrule(lr){3-4} \cmidrule(lr){5-6}
\textbf{\method (Ours)} & Sample-wise Importance
& \textbf{70.26$\pm$1.28} & \textbf{73.90$\pm$0.22} 
& \textbf{46.67$\pm$0.79} & \textbf{49.22$\pm$0.79}   \\
\bottomrule
\\
\end{tabular}
}
\caption{Comparisons to other CL methods with different memory management strategies in the i-Blurry setup for CIFAR10 and CIFAR100 are shown. The i-Blurry-50-10 splits are used for all the datasets and the results are averaged over 3 runs. \method outperforms all other CL methods by large margins on both the $A_\text{AUC}$ and the $A_{avg}$ implying that the sample-wise importance memory management method is effective.} % The best result for each of the metrics is shown in {\bf bold}.}
\label{tab:app_gss}
\end{table}

\rev{
\subsection{Additional Discussion of Other Methods}
\label{sec:comp_to_other_methods}
We present additional discussion on compared methods in Table~\ref{tab:main}.
\subsubsection{Baseline-ER}
ER has been reported to be a simple yet strong method for online CL.~\citep{Prabhu2020GDumbAS, mai2021online} We also observed strong performance for Baseline-ER, as other methods such as EWC++, BiC, and MIR only show marginal to no improvements. However, by using different memory management and memory usage strategy, \method was able to outperform Baseline-ER by large margins
\subsubsection{EWC++}
EWC is a regularization-based method, which regularizes parameters based on the parameter's importance measured by the accumulated Fisher Information. Note that EWC calculates importance of parameters to regularize them, while \method calculates importance of samples to manage episodic memory. We use an online version of EWC, called EWC++, proposed in \citep{rwalk}. EWC++ shows almost no improvement over Baseline-ER, possibly because it was developed as a method to prevent forgetting without using episodic memory. Since episodic memory with ER alleviates most of the forgetting, EWC++ have reduced effect in preventing forgetting, while its side-effect of reducing intransigence still exists~\citep{rwalk}.
\subsubsection{BiC}
BiC uses episodic memory and a distillation loss for training and has an additional bias-correction step to correct the model's bias towards current task's classes. BiC shows strong performance on the disjoint task split. However, it performs poorly for the blurry and the i-Blurry splits, possibly because the bias correction using only two parameters for previous and current tasks cannot correctly capture the bias in the presence of blurry classes.
\subsubsection{ER-MIR}
\label{sec:comp_to_mir}
Note that ER-MIR performs surprisingly similar to the Baseline-ER for CIFAR10, possibly because they are both based on ER. However, ER-MIR does not scale well to larger datasets potentially because its memory usage scheme is not effective for larger datasets.\\\\
To extensively highlight the differences between MIR and \method, we first have to point out that sample memory methods can be discussed in two aspects: memory management and memory usage. MIR uses reservoir sampling for memory management and an improved version of experience replay (ER) for memory usage. MIR improves the memory usage by selecting samples from memory that would suffer the largest loss increase if streamed data were used to train the model (called ‘maximally interfered retrieval’ or MIR).\\\\
MIR can be viewed as using some form of ‘importance weighting’, and has some similarities with our method in the sense that they both utilize the change in loss to determine which samples are used for training. However, these two are different as MIR uses importance scores for selecting samples from a constructed memory whereas CLIB uses it for constructing the memory. Also, MIR assigns high importance scores to samples that show large increases in individual loss when current streamed data are used for training. In contrast, CLIB assigns high importance scores to samples that cause large decreases in memory’s total loss when that sample is used for training. Lastly, MIR’s importance score is temporary, as new importance scores are calculated every iteration, while CLIB accumulates the measured importance scores over time to calculate the overall importance scores.\\\\
Our method performs better than MIR possibly because sample-wise importance memory considers each samples’ importance whereas reservoir sampling does not and memory-only training is not skewed by recently seen samples whereas joint training is. Note that since we are selecting random samples in the memory usage step, a possible future direction would be improving the memory usage as done in MIR.
\subsubsection{GDumb}
GDumb utilizes a greedy sampler, which stores samples greedily, and a dumb learner, which trains a new model from scratch using only the memory whenever an inference query is made.
While we also train only using the memory, we have a completely different memory management method that updates both the model and the memory during training. 
Because GDumb trains from scratch, its performance degrades in the later phases as it cannot accumulate past knowledge.
Not only that, because GDumb has to train from scratch for every inference query, it takes more computation than other methods when there are many inference queries for any-time inference.\\
\subsubsection{Rainbow Memory}
Rainbow memory uses uncertainty based memory management with a two-stage training method that prolongs most of the training until the end of a task. As such, RM performs very poorly on our $A_{AUC}$ metric as it measures any-time inference performance.
Nevertheless, RM performs strongly with respect to the $A_{avg}$ metric which shows the usefulness of our proposed $A_{AUC}$.
}

\begin{table}[t!]
    \centering
    \resizebox{1.0\linewidth}{!}{
\begin{tabular}{ccccccccccccc}
\toprule
\multirow{2}{*}{Methods} & \multicolumn{3}{c}{CIFAR10} & \multicolumn{3}{c}{CIFAR100}            & \multicolumn{3}{c}{TinyImageNet} & \multicolumn{3}{c}{\rev{ImageNet}}\\ \cmidrule(lr){2-4} \cmidrule(lr){5-7} \cmidrule(lr){8-10} \cmidrule(lr){11-13}
     & $A_\text{AUC}$ & $A_\text{avg}$ & \rev{$F_\text{last}$} & $A_\text{AUC}$ & $A_\text{avg}$ & \rev{$F_\text{last}$} & $A_\text{AUC}$ & $A_\text{avg}$ & $\rev{F_\text{last}}$ &
     $\rev{A_\text{AUC}}$ & $\rev{A_\text{avg}}$ & $\rev{F_\text{last}}$\\ \midrule
EWC++           
& 57.34$\pm$2.10 & 60.33$\pm$2.73  & 28.84$\pm$2.35 
& 35.35$\pm$1.96 & 38.78$\pm$2.32  & 32.11$\pm$5.28 
& 22.26$\pm$1.15 & 24.39$\pm$1.18 & 31.65$\pm$1.50
& \rev{24.81} & \rev{26.21} & \rev{44.08}\\
BiC             
& 58.38$\pm$0.54 & 61.49$\pm$0.68  & {22.50$\pm$4.38} 
& 33.51$\pm$3.04 & 37.61$\pm$3.00  & {23.10$\pm$2.87} 
& 22.80$\pm$0.94 & 24.90$\pm$1.07  & 27.01$\pm$0.83
& \rev{27.41} & \rev{28.38} & \rev{36.12}\\
ER-MIR             
& 57.28$\pm$2.43 & 61.93$\pm$3.35  & 27.10$\pm$2.17 
& 35.35$\pm$1.41 & 38.28$\pm$1.15  & 31.76$\pm$4.14 
& 22.10$\pm$1.14 & 24.54$\pm$1.26  & 34.57$\pm$1.39
& \rev{20.48} & \rev{20.68} & \rev{66.96} \\ 
% A-GEM           
% & 39.29$\pm$2.88 & 44.85$\pm$4.70 & 33.12$\pm$5.03 
% & 4.62$\pm$0.23 & 6.94$\pm$0.51 & 15.11$\pm$1.03 
% & 9.42$\pm$2.15 & 12.14$\pm$3.58 & 16.63$\pm$1.39 \\
GDumb           
& 53.20$\pm$1.93 & 55.27$\pm$2.69  & 18.80$\pm$3.40 
& 32.84$\pm$0.45 & 34.03$\pm$0.89  & 22.44$\pm$3.52
& 18.17$\pm$0.19 & 18.69$\pm$0.45 & 17.14$\pm$1.13 
& \rev{14.41} & \rev{14.21} & \rev{31.18} \\
RM\textsuperscript{$\dagger$}
&23.00$\pm$1.43  & 61.52$\pm$3.69 & \textbf{7.01$\pm$2.74}
& 8.63$\pm$0.19  &  33.27$\pm$1.59  & \textbf{7.76$\pm$1.21}
& 5.74$\pm$0.3  & 17.04$\pm$0.77  & \textbf{9.40$\pm$0.40}
& \rev{6.22} & \rev{28.3} & \rev{\textbf{10.71}}\\ 
Baseline-ER   
& 57.46$\pm$2.25 & 60.17$\pm$2.96   & 30.20$\pm$2.96  
& 35.61$\pm$2.08 & 39.10$\pm$2.02  & 31.31$\pm$1.96 
& 22.45$\pm$1.15 & 24.54$\pm$1.26  & 33.02$\pm$1.10
& \rev{25.16} & \rev{26.50} & \rev{44.22}
\\
\cmidrule(lr){1-1} \cmidrule(lr){2-4} \cmidrule(lr){5-7} \cmidrule(lr){8-10} \cmidrule{11-13}
\method (Ours)    
& \textbf{70.26$\pm$1.28} & \textbf{73.90$\pm$0.22} & 23.08$\pm$2.68 
& \textbf{46.67$\pm$0.79} & \textbf{49.22$\pm$0.79} & 23.93$\pm$0.66 
& \textbf{23.87$\pm$0.68} & \textbf{25.05$\pm$0.52} & 23.34$\pm$1.96 
& \rev{\textbf{28.16}} & \rev{\textbf{28.88}} & \rev{38.66}\\
\bottomrule
\\
\end{tabular}
}
\caption{Additional comparisons including the $F_\text{last}$ measure of various online CL methods on the i-Blurry setup for CIFAR10, CIFAR100, TinyImageNet \rev{and ImageNet} are shown. The i-Blurry-50-10 splits are used for all the datasets and the results are averaged over 3 runs \rev{except ImageNet}. Ours outperforms all other CL methods by large margins on both the $A_\text{AUC}$ and the $A_{avg}$. \rev{While \method does not explicitly handle forgetting, it shows similar $F_\text{last}$ with most methods.} The best result for each of the metrics is shown in {\bf bold}.}
\label{tab:main_f}
\end{table}

\begin{figure}[t!]
    \centering
    \resizebox{!}{0.6\linewidth}{
    \includegraphics{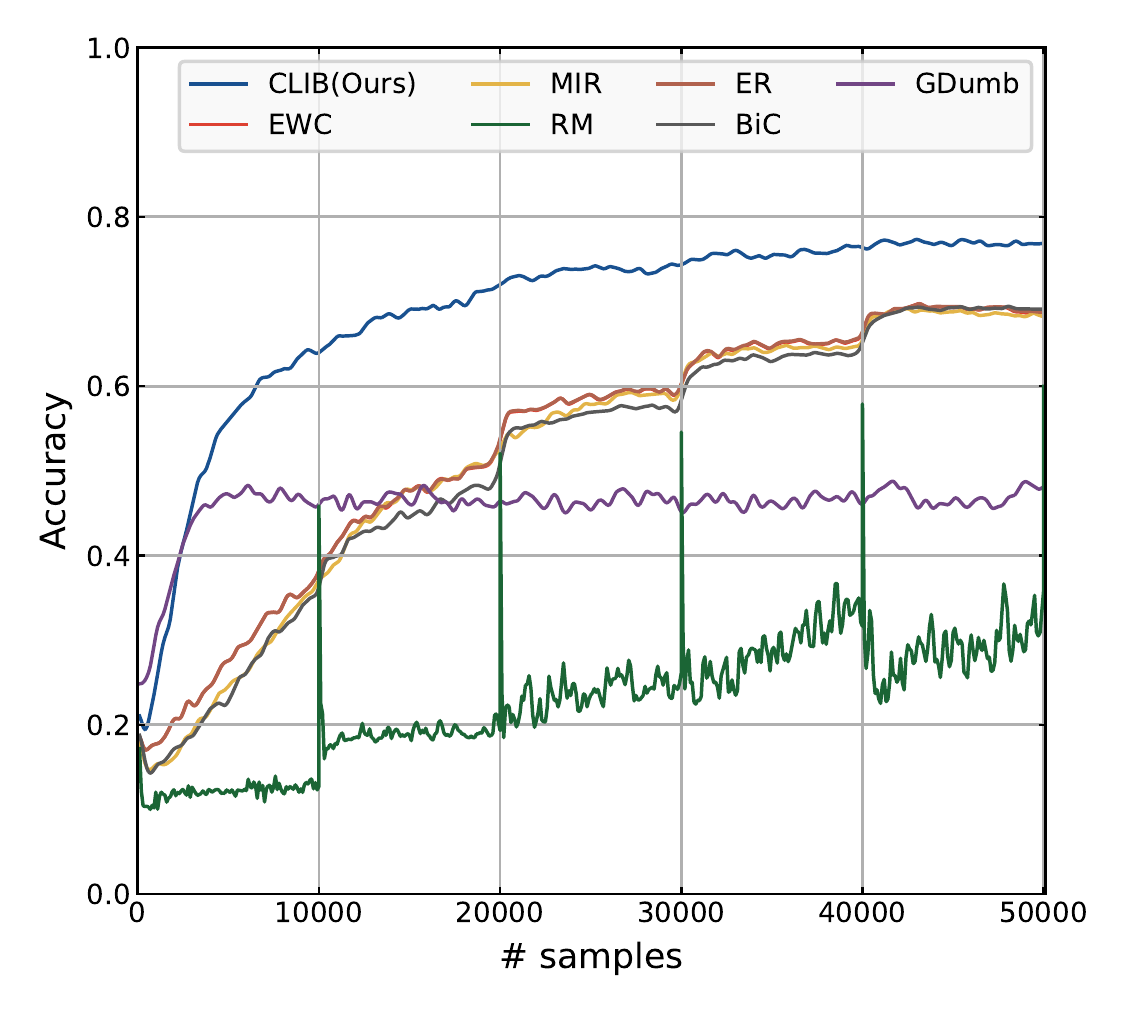}
    }
    \caption{\rev{Accuracy-to-{\# of samples} curve for the blurry setup (no new classes are encountered after the first task). We can see that there is no performance drop for all methods because no new classes are encountered after the first task.}}
    \label{fig:blurry_acc_curve}
\end{figure}

\subsection{Additional Results with the $F_{\text{last}}$ Measure}
\label{sec:app_f_measure}
We report the result of the forgetting measure~\citep{rwalk} here. As in the i-Blurry setup it is not clear which classes belong to each task, we calculate the forgetting class-wise. Note that while forgetting is a useful metric for analyzing stability-plasticity of the method, lower forgetting does not necessarily mean that a CL method is better. For example, if a method do not train with the new task at all, its forgetting will be $0$.

Also, we do not propose a new forgetting measure for anytime inference. It is because forgetting is measured with the best accuracy of each class, and best accuracy usually occur at the end of each task. Thus, measuring the best accuracy among all inferences would not be much different from best accuracy among the inferences at the end of each task.

\rev{Surprisingly, \method shows similar $F_{\text{last}}$ values to most other methods despite not explicitly designed to handle catastrophic forgetting. We believe that the episodic memory used in our memory only training reduces catastrophic forgetting of our method to be similar to other methods.}

\rev{Interestingly, RM shows the best $F_{\text{last}}$ results on all datasets. We believe it is because of RM's two-stage training scheme, which greatly reduces the effect of the streamed data during training. This prevents dominant classes in the task from achieving very high accuracy. Since forgetting is calculated as the difference between the best accuracy and the last accuracy for each class, the forgetting is reduced as the best accuracy itself has been lowered.

Note that GDumb is even less affected by streamed data as it does not train with streamed data at all. However, GDumb does not accumulate learned knowledge, leading to decreasing accuracy as the number of samples per class in the memory decreases. Thus, while GDumb scores relatively good in forgetting considering the decreasing accuracy trend, it ultimately scores worse in forgetting than RM.}

\rev{
\subsection{Additional Discussion on Potential Performance Drop at Later Phases}
\label{sec:perf_drop}
We discuss the potential cause for the `performance drop' of \method in the i-Blurry setup as is shown in Fig.~\ref{fig:iblurry_plot}. We argue that the `performance drop' is attributed to the number of encountered classes increasing in i-Blurry setups at later phases. 
To see what would happen when the number of encountered classes do not increase at later phases, we show the accuracy-to-{\# of samples} curve for the blurry setup results in Table~\ref{tab:class_dist} for $N=0$ in Fig.~\ref{fig:blurry_acc_curve}, where no new classes are encountered after the first task. There is no apparent performance drop at later phases for all methods, which supports our argument that the drop is due to more and more classes being encountered at later phases.}

\subsection{Performance of Exponential with Reset LR Schedule}
We show brief results for the LR schedule used in our baseline in Table~\ref{tab:exp_reset}.
We compare the constant LR with the exponential with reset used in our baseline.
The exponential with reset is better than the constant LR, which is why we used it in our baseline.

\begin{table}[t!]
    \centering
    \resizebox{0.6\linewidth}{!}{
\begin{tabular}{cccc}
\toprule
Methods & LR Schedule & $A_\text{AUC}$ & $A_\text{avg}$\\ \midrule
\multirow{2}{*}{Baseline-ER} & Constant           
& 56.73$\pm$1.73 & 58.18$\pm$3.41\\
& Exp w/ Reset 
& 57.46$\pm$2.25 & 60.17$\pm$2.96 \\
\midrule
\multirow{2}{*}{EWC++} & Constant          
& 56.54$\pm$1.57 & 57.92$\pm$3.26 \\
& Exp w/ Reset 
& 57.34$\pm$2.10 & 60.33$\pm$2.73 \\
\bottomrule
\\
\end{tabular}
}
\caption{Comparison between exponential with reset schedule and constant LR on CIFAR10 are shown. It shows that our baseline LR schedule, exponential with reset, is reasonable. It shows better performance than constant LR, especially in $A_\text{avg}$ metric.}
\label{tab:exp_reset}
\end{table}

%% file: ms.bbl
\begin{thebibliography}{51}
\providecommand{\natexlab}[1]{#1}
\providecommand{\url}[1]{\texttt{#1}}
\expandafter\ifx\csname urlstyle\endcsname\relax
  \providecommand{\doi}[1]{doi: #1}\else
  \providecommand{\doi}{doi: \begingroup \urlstyle{rm}\Url}\fi

\bibitem[Aljundi et~al.(2017)Aljundi, Chakravarty, and
  Tuytelaars]{aljundi17expert}
R.~Aljundi, P.~Chakravarty, and T~Tuytelaars.
\newblock {Expert Gate: Lifelong Learning with a Network of Experts}.
\newblock In \emph{CVPR}, 2017.

\bibitem[Aljundi et~al.(2018)Aljundi, Babiloni, Elhoseiny, Rohrbach, and
  Tuytelaars]{Aljundi2018MemoryAS}
Rahaf Aljundi, Francesca Babiloni, Mohamed Elhoseiny, Marcus Rohrbach, and
  Tinne Tuytelaars.
\newblock Memory aware synapses: Learning what (not) to forget.
\newblock In \emph{ECCV}, 2018.

\bibitem[Aljundi et~al.(2019{\natexlab{a}})Aljundi, Belilovsky, Tuytelaars,
  Charlin, Caccia, Lin, and Page-Caccia]{aljundi2019online}
Rahaf Aljundi, Eugene Belilovsky, Tinne Tuytelaars, Laurent Charlin, Massimo
  Caccia, Min Lin, and Lucas Page-Caccia.
\newblock Online continual learning with maximal interfered retrieval.
\newblock \emph{Advances in Neural Information Processing Systems},
  32:\penalty0 11849--11860, 2019{\natexlab{a}}.

\bibitem[Aljundi et~al.(2019{\natexlab{b}})Aljundi, Kelchtermans, and
  Tuytelaars]{aljundi2019task}
Rahaf Aljundi, Klaas Kelchtermans, and Tinne Tuytelaars.
\newblock Task-free continual learning.
\newblock In \emph{Proceedings of the IEEE/CVF Conference on Computer Vision
  and Pattern Recognition}, pp.\  11254--11263, 2019{\natexlab{b}}.

\bibitem[Aljundi et~al.(2019{\natexlab{c}})Aljundi, Lin, Goujaud, and
  Bengio]{aljundi2019gradient}
Rahaf Aljundi, Min Lin, Baptiste Goujaud, and Yoshua Bengio.
\newblock Gradient based sample selection for online continual learning.
\newblock In \emph{NeurIPS}, pp.\  11816--11825, 2019{\natexlab{c}}.

\bibitem[Bang et~al.(2021)Bang, Kim, Yoo, Ha, and Choi]{bang2021rainbow}
Jihwan Bang, Heesu Kim, YoungJoon Yoo, Jung-Woo Ha, and Jonghyun Choi.
\newblock Rainbow memory: Continual learning with a memory of diverse samples.
\newblock In \emph{Proceedings of the IEEE/CVF Conference on Computer Vision
  and Pattern Recognition}, pp.\  8218--8227, 2021.

\bibitem[Castro et~al.(2018)Castro, Marin-Jimenez, Guil, Schmid, and
  Alahari]{castro2018eccv}
Francisco~M. Castro, Manuel~J. Marin-Jimenez, Nicolas Guil, Cordelia Schmid,
  and Karteek Alahari.
\newblock End-to-end incremental learning.
\newblock In \emph{ECCV}, 2018.

\bibitem[Chang et~al.(2017)Chang, Learned-Miller, and
  McCallum]{chang2017active}
Haw-Shiuan Chang, Erik Learned-Miller, and Andrew McCallum.
\newblock Active bias: Training more accurate neural networks by emphasizing
  high variance samples.
\newblock \emph{Advances in Neural Information Processing Systems},
  30:\penalty0 1002--1012, 2017.

\bibitem[Chaudhry et~al.(2018{\natexlab{a}})Chaudhry, Dokania, Ajanthan, and
  Torr]{Chaudhry_2018_ECCV}
Arslan Chaudhry, Puneet~K. Dokania, Thalaiyasingam Ajanthan, and Philip H.~S.
  Torr.
\newblock Riemannian walk for incremental learning: Understanding forgetting
  and intransigence.
\newblock In \emph{ECCV}, 2018{\natexlab{a}}.

\bibitem[Chaudhry et~al.(2018{\natexlab{b}})Chaudhry, Dokania, Ajanthan, and
  Torr]{rwalk}
Arslan Chaudhry, Puneet~K. Dokania, Thalaiyasingam Ajanthan, and Philip H.~S.
  Torr.
\newblock Riemannian walk for incremental learning: Understanding forgetting
  and intransigence.
\newblock In \emph{ECCV}, 2018{\natexlab{b}}.

\bibitem[Chaudhry et~al.(2019)Chaudhry, Ranzato, Rohrbach, and Elhoseiny]{AGEM}
Arslan Chaudhry, Marc’Aurelio Ranzato, Marcus Rohrbach, and Mohamed
  Elhoseiny.
\newblock Efficient lifelong learning with {A-GEM}.
\newblock In \emph{ICLR}, 2019.

\bibitem[Cheung et~al.(2019)Cheung, Terekhov, Chen, Agrawal, and
  Olshausen]{cheung19superposition}
B.~Cheung, A.~Terekhov, Y.~Chen, P.~Agrawal, and B.~Olshausen.
\newblock {Superposition of Many Models into One}.
\newblock In \emph{NeurIPS}, 2019.

\bibitem[Cong et~al.(2020)Cong, Zhao, Li, Wang, and Carin]{Cong2020GANMW}
Yulai Cong, Miaoyun Zhao, J.~Li, Sijia Wang, and L.~Carin.
\newblock {GAN} memory with no forgetting.
\newblock In \emph{NeurIPS}, 2020.

\bibitem[Csiba \& Richt{\'a}rik(2018)Csiba and
  Richt{\'a}rik]{csiba2018importance}
Dominik Csiba and Peter Richt{\'a}rik.
\newblock Importance sampling for minibatches.
\newblock \emph{The Journal of Machine Learning Research}, 19\penalty0
  (1):\penalty0 962--982, 2018.

\bibitem[Cubuk et~al.(2019)Cubuk, Zoph, Mane, Vasudevan, and Le]{AutoAugment}
Ekin~D. Cubuk, Barret Zoph, Dandelion Mane, Vijay Vasudevan, and Quoc~V. Le.
\newblock {AutoAugment}: Learning augmentation strategies from data.
\newblock In \emph{CVPR}, June 2019.

\bibitem[Ebrahimi et~al.(2020)Ebrahimi, Elhoseiny, Darrell, and
  Rohrbach]{ebrahimi20uncertainty}
S.~Ebrahimi, M.~Elhoseiny, T.~Darrell, and M.~Rohrbach.
\newblock {Uncertainty-Guided Continual Learning with Bayesian Neural
  Networks}.
\newblock In \emph{ICLR}, 2020.

\bibitem[He et~al.(2020)He, Mao, Shao, and Zhu]{he20incremental}
J.~He, R.~Mao, Z.~Shao, and F.~Zhu.
\newblock {Incremental Learning In Online Scenario}.
\newblock In \emph{CVPR}, 2020.

\bibitem[Hu et~al.(2019)Hu, Lin, Liu, Tao, Tao, Ma, Zhao, and
  Yan]{hu2018overcoming}
Wenpeng Hu, Zhou Lin, Bing Liu, Chongyang Tao, Zhengwei Tao, Jinwen Ma, Dongyan
  Zhao, and Rui Yan.
\newblock Overcoming catastrophic forgetting via model adaptation.
\newblock In \emph{ICLR}, 2019.

\bibitem[Jin et~al.(2020)Jin, Du, and Ren]{gmed}
Xisen Jin, Junyi Du, and Xiang Ren.
\newblock Gradient based memory editing for task-free continual learning.
\newblock \emph{arXiv preprint arXiv:2006.15294}, 2020.

\bibitem[Katharopoulos \& Fleuret(2018)Katharopoulos and
  Fleuret]{katharopoulos2018not}
Angelos Katharopoulos and Fran{\c{c}}ois Fleuret.
\newblock Not all samples are created equal: Deep learning with importance
  sampling.
\newblock In \emph{International conference on machine learning}, pp.\
  2525--2534. PMLR, 2018.

\bibitem[Kim et~al.(2020)Kim, Jeong, and Kim]{kim20PRS}
C.~Kim, J.~Jeong, and G.~Kim.
\newblock {Imbalanced Continual Learning with Partitioning Reservoir Sampling}.
\newblock In \emph{ECCV}, 2020.

\bibitem[Kim et~al.(2019)Kim, Bae, Jo, and Choi]{kimmer}
Dahyun Kim, Jihwan Bae, Yeonsik Jo, and Jonghyun Choi.
\newblock Incremental learning with maximum entropy regularization: Rethinking
  forgetting and intransigence.
\newblock \emph{arXiv preprint https://arxiv.org/abs/1902.00829}, 2019.

\bibitem[Kim et~al.(2018)Kim, Kim, Seo, Kim, Park, Park, Jo, Kim, Yang, Kim,
  et~al.]{kim2018nsml}
Hanjoo Kim, Minkyu Kim, Dongjoo Seo, Jinwoong Kim, Heungseok Park, Soeun Park,
  Hyunwoo Jo, KyungHyun Kim, Youngil Yang, Youngkwan Kim, et~al.
\newblock Nsml: Meet the mlaas platform with a real-world case study.
\newblock \emph{arXiv preprint arXiv:1810.09957}, 2018.

\bibitem[Kirkpatrick et~al.(2017)Kirkpatrick, Pascanu, Rabinowitz, Veness,
  Desjardins, Rusu, Milan, Quan, Ramalho, Grabska-Barwinska, et~al.]{ewc}
James Kirkpatrick, Razvan Pascanu, Neil Rabinowitz, Joel Veness, Guillaume
  Desjardins, Andrei~A Rusu, Kieran Milan, John Quan, Tiago Ramalho, Agnieszka
  Grabska-Barwinska, et~al.
\newblock Overcoming catastrophic forgetting in neural networks.
\newblock \emph{Proceedings of the national academy of sciences}, 114\penalty0
  (13):\penalty0 3521--3526, 2017.

\bibitem[Kloek \& Van~Dijk(1978)Kloek and Van~Dijk]{kloek1978bayesian}
Tuen Kloek and Herman~K Van~Dijk.
\newblock Bayesian estimates of equation system parameters: an application of
  integration by monte carlo.
\newblock \emph{Econometrica: Journal of the Econometric Society}, pp.\  1--19,
  1978.

\bibitem[LeCun et~al.(1990)LeCun, Denker, and Solla]{lecun1990optimal}
Yann LeCun, John~S Denker, and Sara~A Solla.
\newblock Optimal brain damage.
\newblock In \emph{Advances in neural information processing systems}, pp.\
  598--605, 1990.

\bibitem[Lee et~al.(2017{\natexlab{a}})Lee, Yoon, Yang, and
  Hwang]{Lee2017LifelongLW}
Jeongtae Lee, Jaehong Yoon, Eunho Yang, and Sung~Ju Hwang.
\newblock Lifelong learning with dynamically expandable networks.
\newblock \emph{ICLR}, 2017{\natexlab{a}}.

\bibitem[Lee et~al.(2017{\natexlab{b}})Lee, Kim, Ha, and
  Zhang]{Lee2017OvercomingCF}
Sang-Woo Lee, Jin-Hwa Kim, JungWoo Ha, and Byoung-Tak Zhang.
\newblock Overcoming catastrophic forgetting by incremental moment matching.
\newblock In \emph{NeurIPS}, 2017{\natexlab{b}}.

\bibitem[Lee et~al.(2019)Lee, Ha, Zhang, and Kim]{lee2019neural}
Soochan Lee, Junsoo Ha, Dongsu Zhang, and Gunhee Kim.
\newblock A neural dirichlet process mixture model for task-free continual
  learning.
\newblock In \emph{International Conference on Learning Representations}, 2019.

\bibitem[Liu et~al.(2020)Liu, Su, Liu, Schiele, and Sun]{mnemonics}
Yaoyao Liu, Yuting Su, An-An Liu, Bernt Schiele, and Qianru Sun.
\newblock Mnemonics training: Multi-class incremental learning without
  forgetting.
\newblock In \emph{Proceedings of the IEEE/CVF Conference on Computer Vision
  and Pattern Recognition}, pp.\  12245--12254, 2020.

\bibitem[Lopez-Paz \& Ranzato(2017)Lopez-Paz and
  Ranzato]{LopezPaz2017GradientEM}
David Lopez-Paz and Marc'Aurelio Ranzato.
\newblock Gradient episodic memory for continual learning.
\newblock In \emph{NIPS}, 2017.

\bibitem[Losing et~al.(2018)Losing, Hammer, and Wersing]{losing18incremental}
Viktor Losing, Barbara Hammer, and Heiko Wersing.
\newblock Incremental on-line learning: A review and comparison of state of the
  art algorithms.
\newblock \emph{Neurocomputing}, 275:\penalty0 1261 -- 1274, 2018.
\newblock ISSN 0925-2312.

\bibitem[Mai et~al.(2021)Mai, Li, Jeong, Quispe, Kim, and
  Sanner]{mai2021online}
Zheda Mai, Ruiwen Li, Jihwan Jeong, David Quispe, Hyunwoo Kim, and Scott
  Sanner.
\newblock Online continual learning in image classification: An empirical
  survey.
\newblock \emph{arXiv preprint arXiv:2101.10423}, 2021.

\bibitem[Mallya \& Lazebnik(2018)Mallya and Lazebnik]{mallya17packnet}
A.~Mallya and S.~Lazebnik.
\newblock {PackNet: Adding Multiple Tasks to a Single Network by Iterative
  Pruning}.
\newblock In \emph{CVPR}, 2018.

\bibitem[McCloskey \& Neal(1989)McCloskey and Neal]{mccloskeyC89}
M.~McCloskey and Neal.
\newblock Catastrophic interference in connectionist networks: The sequential
  learning problem.
\newblock \emph{Psychology of Learning and Motivation}, 24:\penalty0 109--165,
  1989.

\bibitem[Mirzadeh et~al.(2020)Mirzadeh, Farajtabar, Pascanu, and
  Ghasemzadeh]{mirzadeh2020understanding}
Seyed~Iman Mirzadeh, Mehrdad Farajtabar, Razvan Pascanu, and Hassan
  Ghasemzadeh.
\newblock Understanding the role of training regimes in continual learning.
\newblock In \emph{Advances in Neural Information Processing Systems 33: Annual
  Conference on Neural Information Processing Systems 2020}, 2020.

\bibitem[Prabhu et~al.(2020)Prabhu, Torr, and Dokania]{Prabhu2020GDumbAS}
Ameya Prabhu, P.~Torr, and Puneet~K. Dokania.
\newblock {GDumb}: A simple approach that questions our progress in continual
  learning.
\newblock In \emph{ECCV}, 2020.

\bibitem[Ratcliff(1990)]{ratcliff90}
R.~Ratcliff.
\newblock {Connectionist models of recognition memory: Constraints imposed by
  learning and forgetting functions}.
\newblock \emph{Psychological Review}, 97\penalty0 (2):\penalty0 285--308,
  1990.

\bibitem[Rebuffi et~al.(2017)Rebuffi, Kolesnikov, Sperl, and Lampert]{icarl}
Sylvestre-Alvise Rebuffi, Alexander Kolesnikov, Georg Sperl, and Christoph~H.
  Lampert.
\newblock {iCaRL}: Incremental classifier and representation learning.
\newblock In \emph{CVPR}, 2017.

\bibitem[Rolnick et~al.(2018)Rolnick, Ahuja, Schwarz, Lillicrap, and Wayne]{ER}
David Rolnick, Arun Ahuja, Jonathan Schwarz, Timothy~P Lillicrap, and Greg
  Wayne.
\newblock Experience replay for continual learning.
\newblock \emph{arXiv preprint arXiv:1811.11682}, 2018.

\bibitem[Rusu et~al.(2016)Rusu, Rabinowitz, Desjardins, Soyer, Kirkpatrick,
  Kavukcuoglu, Pascanu, and Hadsell]{Rusu2016ProgressiveNN}
Andrei~A. Rusu, Neil~C. Rabinowitz, Guillaume Desjardins, Hubert Soyer, James
  Kirkpatrick, Koray Kavukcuoglu, Razvan Pascanu, and Raia Hadsell.
\newblock Progressive neural networks.
\newblock \emph{arXiv}, abs/1606.04671, 2016.

\bibitem[Saha et~al.(2021)Saha, Garg, and Roy]{saha2021gradient}
Gobinda Saha, Isha Garg, and Kaushik Roy.
\newblock Gradient projection memory for continual learning.
\newblock In \emph{International Conference on Learning Representations}, 2021.

\bibitem[Shin et~al.(2017)Shin, Lee, Kim, and Kim]{Shin2017ContinualLW}
Hanul Shin, Jung~Kwon Lee, Jaehong Kim, and Jiwon Kim.
\newblock Continual learning with deep generative replay.
\newblock In \emph{NeurIPS}, 2017.

\bibitem[Sung et~al.(2017)Sung, Kim, Jo, Yang, Kim, Lausen, Kim, Lee, Kwak, Ha,
  et~al.]{sung2017nsml}
Nako Sung, Minkyu Kim, Hyunwoo Jo, Youngil Yang, Jingwoong Kim, Leonard Lausen,
  Youngkwan Kim, Gayoung Lee, Donghyun Kwak, Jung-Woo Ha, et~al.
\newblock Nsml: A machine learning platform that enables you to focus on your
  models.
\newblock \emph{arXiv preprint arXiv:1712.05902}, 2017.

\bibitem[van~de Ven et~al.(2021)van~de Ven, Li, and Tolias]{van2021class}
Gido~M van~de Ven, Zhe Li, and Andreas~S Tolias.
\newblock Class-incremental learning with generative classifiers.
\newblock In \emph{CVPR}, 2021.

\bibitem[Vitter(1985)]{reservoir}
Jeffrey~S Vitter.
\newblock Random sampling with a reservoir.
\newblock \emph{ACM Transactions on Mathematical Software (TOMS)}, 11\penalty0
  (1):\penalty0 37--57, 1985.

\bibitem[Wu et~al.(2018)Wu, Herranz, Liu, Wang, van~de Weijer, and
  Raducanu]{Wu2018MemoryRG}
Chenshen Wu, Luis Herranz, Xialei Liu, Yaxing Wang, Joost van~de Weijer, and
  Bogdan Raducanu.
\newblock {Memory Replay GANs}: learning to generate images from new categories
  without forgetting.
\newblock In \emph{NeurIPS}, 2018.

\bibitem[Wu et~al.(2019)Wu, Chen, Wang, Ye, Liu, Guo, and Fu]{bic}
Yue Wu, Yan-Jia Chen, Lijuan Wang, Yuancheng Ye, Zicheng Liu, Yandong Guo, and
  Yun Fu.
\newblock Large scale incremental learning.
\newblock In \emph{CVPR}, 2019.

\bibitem[Yoon et~al.(2020)Yoon, Kim, Yang, and Hwang]{yoon2020apd}
Jaehong Yoon, Saehoon Kim, Eunho Yang, and Sung~Ju Hwang.
\newblock Scalable and order-robust continual learning with additive parameter
  decomposition.
\newblock In \emph{ICLR}, 2020.

\bibitem[Yun et~al.(2019)Yun, Han, Oh, Chun, Choe, and Yoo]{yun2019cutmix}
Sangdoo Yun, Dongyoon Han, Seong~Joon Oh, Sanghyuk Chun, Junsuk Choe, and
  Youngjoon Yoo.
\newblock Cutmix: Regularization strategy to train strong classifiers with
  localizable features.
\newblock In \emph{ICCV}, pp.\  6023--6032, 2019.

\bibitem[Zenke et~al.(2017)Zenke, Poole, and Ganguli]{Zenke2017ContinualLT}
Friedemann Zenke, Ben Poole, and Surya Ganguli.
\newblock Continual learning through synaptic intelligence.
\newblock In \emph{ICML}, 2017.

\end{thebibliography}
